\documentclass[10pt,journal,compsoc]{IEEEtran}
\ifCLASSOPTIONcompsoc
  \usepackage[nocompress]{cite}
\else
  \usepackage{cite}
\fi
\ifCLASSINFOpdf
\else
\fi

\usepackage{graphicx}
\usepackage{amsmath}
\usepackage{amssymb}
\usepackage{booktabs}
\usepackage{xcolor}

\usepackage{ulem}
\makeatletter
\renewcommand{\maketag@@@}[1]{\hbox{\m@th\normalsize\normalfont#1}}%
\makeatother
\usepackage{algorithm}  
\usepackage{algorithmic} 
\usepackage[accsupp]{axessibility}
\usepackage{ragged2e}
\usepackage{bbding}
\usepackage{multirow}
\usepackage{subcaption}
\begin{document}
%
\title{Stimulative Training++: Go Beyond The Performance Limits of Residual Networks}

\author{
Peng Ye, Tong He, Shengji Tang, Baopu Li, Tao Chen, Lei Bai, Wanli Ouyang
\IEEEcompsocitemizethanks{\IEEEcompsocthanksitem Peng Ye, Shengji Tang and Tao Chen are with the School of Information Science and Technology, Fudan University, Shanghai, China \\
E-mail: \{20110720039, eetchen\}@fudan.edu.cn
\IEEEcompsocthanksitem Tong He, Lei Bai and Wanli Ouyang are with the Shanghai AI Laboratory, Shanghai, China.
\IEEEcompsocthanksitem Baopu Li is with the Oracle, USA.
\IEEEcompsocthanksitem Work is partially performed when Peng Ye is an intern at Shanghai AI Laboratory, China. Tao Chen is the corresponding author.}
}

\IEEEtitleabstractindextext{%
\begin{abstract}\justifying
Residual networks have shown great success and become indispensable in recent deep neural network models. 
In this work, we aim to re-investigate the training process of residual networks from a novel social psychology perspective of loafing, and further propose a new training scheme as well as three improved strategies for boosting residual networks beyond their performance limits.
Previous research has suggested that residual networks can be considered as ensembles of shallow networks, which implies that the final performance of a residual network is influenced by a group of subnetworks.
We identify a previously overlooked problem that is analogous to social loafing, where subnetworks within a residual network are prone to exert less effort when working as part of a group compared to working alone.
We define this problem as \textit{network loafing}. 
Similar to the decreased individual productivity and overall performance as demonstrated in society, network loafing inevitably causes sub-par performance. Inspired by solutions from social psychology, we first propose a novel training scheme called stimulative training,
which randomly samples a residual subnetwork and calculates the KL divergence loss between the sampled subnetwork and the given residual network for extra supervision.
In order to unleash the potential of stimulative training, we further propose three simple-yet-effective strategies, including a novel KL- loss 
that only aligns the network logits direction,
random smaller inputs for subnetworks, and inter-stage sampling rules.
Comprehensive experiments and analysis verify the effectiveness of stimulative training as well as its three improved strategies. 
For example, the proposed method can boost the performance of ResNet50 on ImageNet to 80.5\% Top1 accuracy without using any extra data, model, trick, or changing the structure. With only uniform augment, the performance can be further improved to 81.0\% Top1 accuracy, better than the best training recipes provided by Timm library and PyTorch official version. We also verify its superiority on various typical models, datasets, and tasks and give some theoretical analysis. As such, we advocate utilizing the proposed method as a general and next-generation technology to train residual networks. The code is available at 
{https://github.com/Sunshine-Ye/NIPS22-ST}.

\end{abstract}

}

\maketitle

\IEEEdisplaynontitleabstractindextext

%
\IEEEpeerreviewmaketitle

\IEEEraisesectionheading{\section{Introduction}\label{sec:introduction}}
\IEEEPARstart{S}{ince} ResNet~\cite{he2016deep} wins the first place at the ILSVRC-2015 competition, simple-yet-effective residual connections are applied in various deep networks, such as CNN, MLP, and transformer. To explore the secrets behind the success of residual networks, numerous studies have been proposed. 
He et. al~\cite{he2016deep} first exploit the residual structure to avoid performance degradation of deep networks. Further, 
He et. al~\cite{he2016identity} consider that such a structure can transfer low-level features to high-level layers in forward propagation and directly transmit the gradients from deep to shallow layers in backward propagation. Balduzzi et. al~\cite{balduzzi2017shattered} find that residual networks can alleviate the shattered gradients problem that gradients resemble white noise. In addition, Veit et. al~\cite{veit2016residual} experimentally verify that residual networks can be seen as a collection of numerous networks of different lengths, namely \textit{unraveled view}. Following this view, Sun et. al~\cite{sun2022low} further attribute the success of residual networks to shallow subnetworks, which may correspond to the low-degree term when regarding the neural network as a polynomial function. The authors in ~\cite{barzilai2022kernel} theoretically prove that the eigenvalues of the residual convolutional neural tangent kernel (CNTK) consist of weighted sums of eigenvalues of CNTK of subnetworks. Since the unraveled view is supported both experimentally and theoretically, we further investigate some interesting mechanisms behind residual networks based on this inspiring work. Specifically, we treat a residual network as an ensemble of relatively shallow subnetworks and consider its final performance to be co-determined by a group of subnetworks.
\begin{figure*}[t]
  \centering
  \subcaptionbox{ResNet on CIFAR10}{
    \label{fig:loafing-resnet} 
    \includegraphics[width=0.32\linewidth]{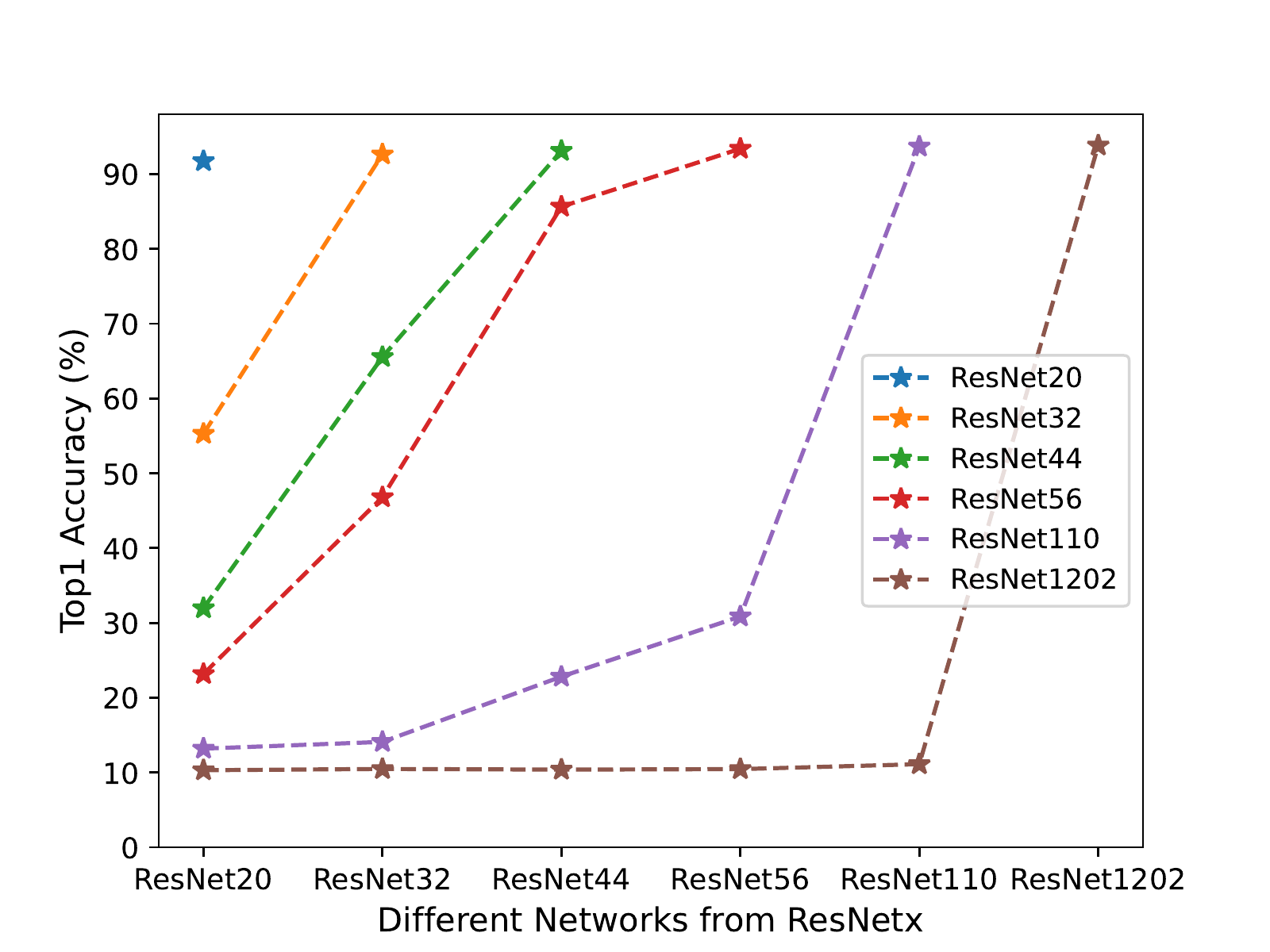}}
  \subcaptionbox{DenseNet on CIFAR100}{
    \label{fig:loafing-densenet} 
    \includegraphics[width=0.32\linewidth]{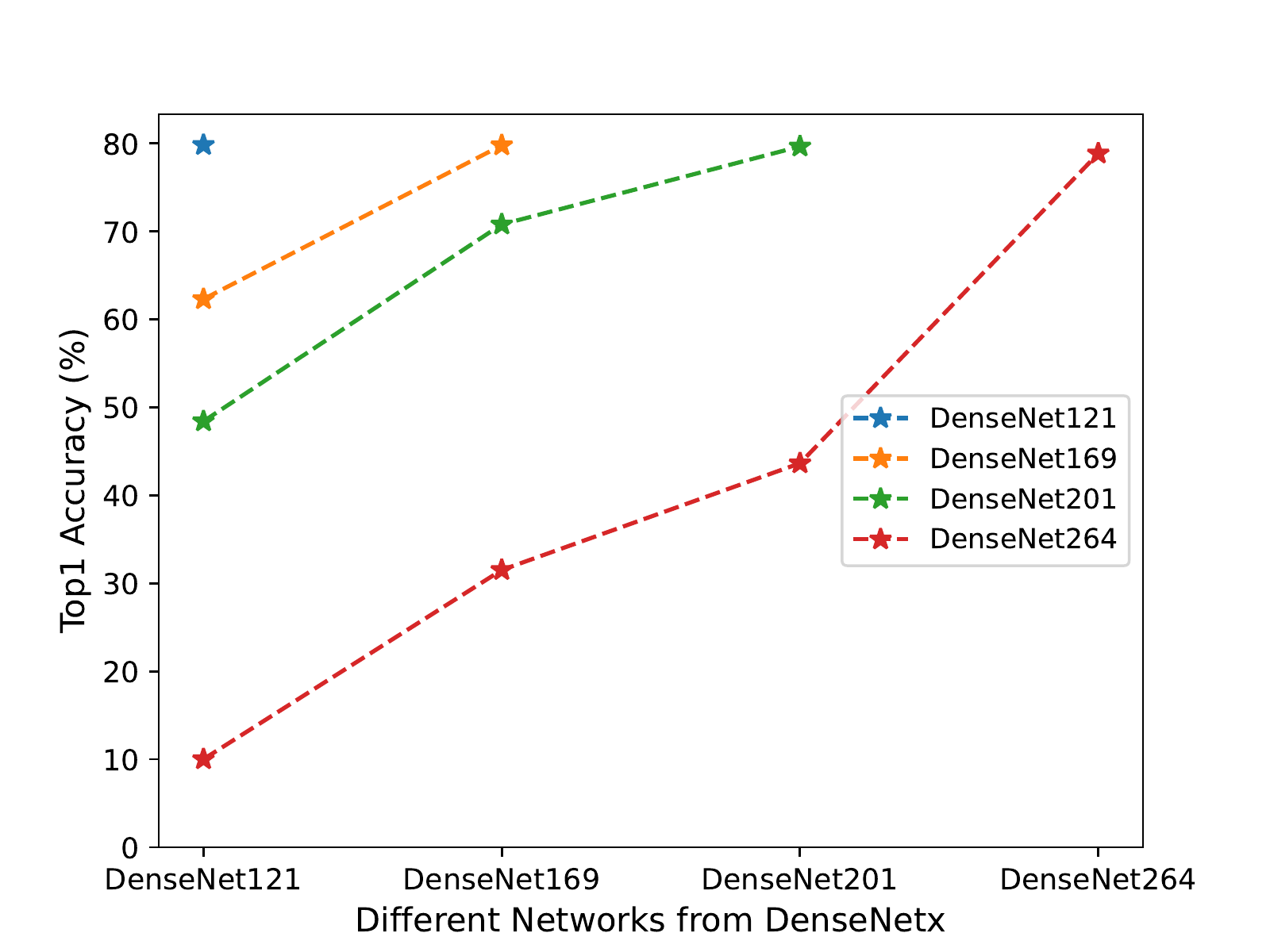}}
  \subcaptionbox{Transformer on ImageNet}{
    \label{fig:loafing-vit} 
    \includegraphics[width=0.32\linewidth]{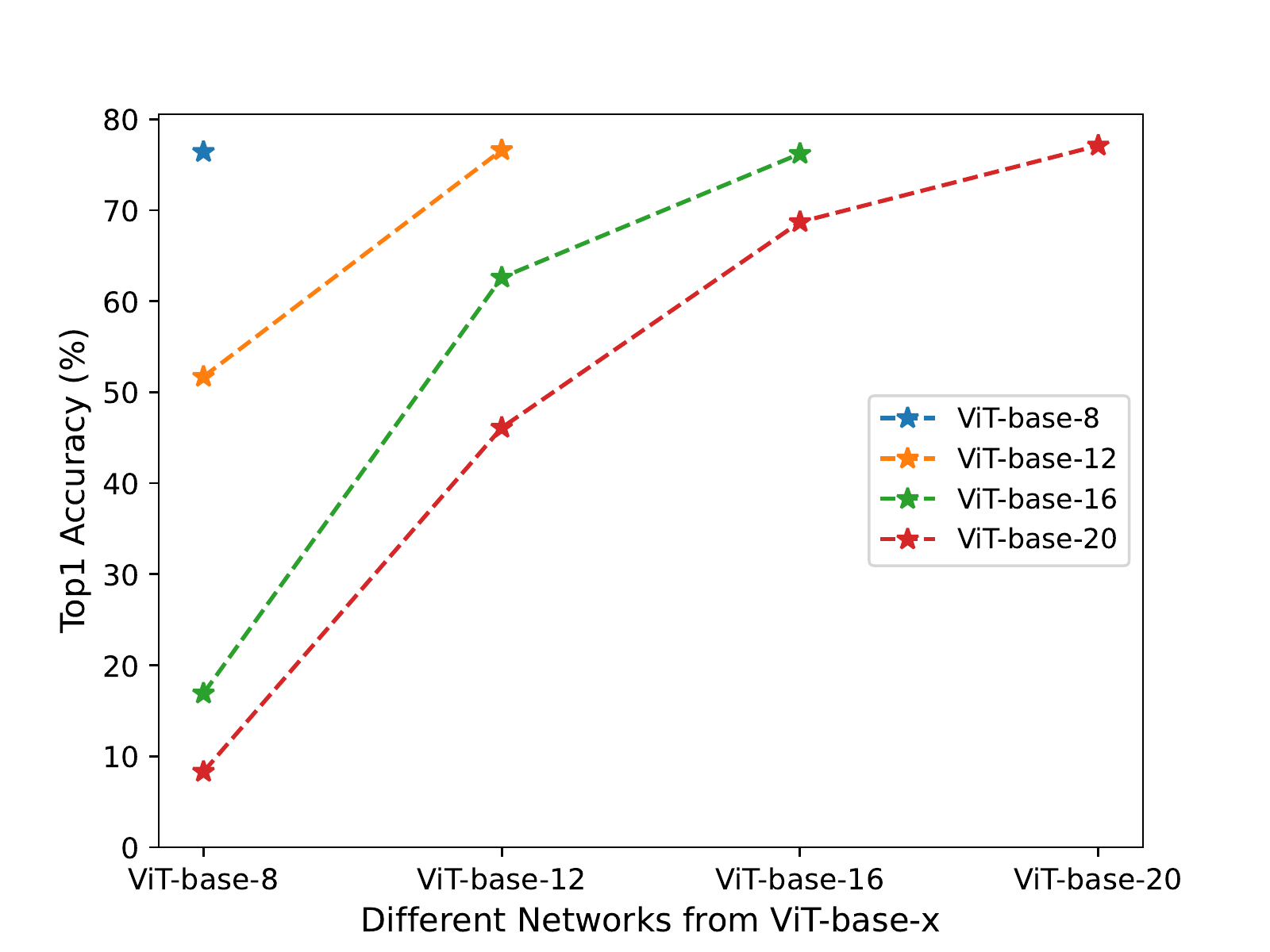}}
  \vspace{-2mm}
  \caption{Different residual networks invariably suffer from the problem of network loafing, and deeper residual network tends to have a more serious loafing problem. Such a problem widely exists in kinds of residual architectures and datasets of different sizes.
  The horizontal axis means the sampled different subnetworks from ResNetx, DenseNetx, or ViTx.}
  \label{fig:loafing_diff_networks} 
\end{figure*}
\par In social psychology, working in a group is always a tricky thing.
Compared with performing tasks alone, group members tend to make less effort when working as part of a group, which is defined as the social loafing problem~\cite{ingham1974ringelmann,petty1977effects}. 
Moreover, social psychology researches indicate that increasing the group size may aggravate social loafing for the decrease of individual visibility~\cite{williams1991social,george1995asymmetrical}.
Interestingly, a similar problem and behavior can also be observed in ensemble-like residual networks. As shown in Fig.~\ref{fig:loafing_diff_networks}, different residual networks invariably suffer from the loafing problem, that is, the subnetworks working in a given residual network are prone to exert degraded performances than these subnetworks working individually. 
For example, selecting a subnetwork of ResNet32 from ResNet56 only has a Top1 accuracy of 46.80\%, which is much lower than the ResNet32 trained individually with an accuracy of 92.63\%. Moreover, the loafing problem in deeper residual networks is more severe than shallower ones, that is, the same subnetwork in deeper residual networks constantly presents inferior performance than that in shallower residual networks. For example, ResNet20 within ResNet32 has a Top1 accuracy of 55.28\%, while the ResNet20 within ResNet56 (deeper than ResNet 32) only has a 23.17\% Top1 accuracy. To the best of our knowledge, such problems have not been addressed in the literature. Hereafter, we define this previously overlooked problem as \textit{network loafing}.\footnote{\textit{Network loafing} 
is a metaphorical description used to characterize a behavior observed in neural networks, which is not directly linked to biological processes.} As social psychology researches have shown that social loafing will ultimately cause low productivity of each individual and the collective~\cite{latane1979many,simms2014social},
we consider that network loafing may also hinder the performance of a given residual network and its subnetworks.
\par In social psychology, there exist two commonly-used solutions for preventing the social loafing problem within groups: (1) establishing individual accountability by increasing individual supervision; (2) making tasks cooperative by setting up consistent overall goal~\cite{williams1991social,george1995asymmetrical}. Inspired by this, we propose a novel training strategy for improving residual networks, namely \textit{stimulative training}. In detail, for each mini-batch during stimulative training, besides the main loss of the given residual network in conventional training, we will randomly sample a residual subnetwork (individual supervision) and calculate the KL divergence loss between the sampled subnetwork and the given residual network (consistent overall goal). 
The proposed stimulative training can effectively address the problem of network loafing by enhancing the individual performance of sub-networks and aligning their goals with the given residual network, ultimately leading to enhanced performance during training.
\par To further unleash the potential of stimulative training, we improve it with three simple-yet-effective strategies, namely \textit{stimulative training++}. First, we propose a novel KL- loss to replace the naive KL loss, which eliminates the possible negative influence of the logit magnitude and only aligns the logit direction between the main network and diverse subnetworks. Second, we provide sampled subnetworks with random smaller inputs instead of the same input resolution of the main network, enabling sampled subnetworks with multi-scale and larger effective receptive fields. Third, we investigate the popular residual network families and conclude inter-stage sampling rules for better designing a sampling space. We show that despite their simplicity, all the above three strategies can significantly improve the results of stimulative training. Moreover, we conduct comprehensive experiments and analyses to demonstrate why and how the above three strategies work. 
\par Without bells and whistles, stimulative training++ can boost residual networks beyond their performance limits. Specifically, by replacing common training with the proposed stimulative training++, the performance of ResNet50 on ImageNet can be improved from a Top1 accuracy of 76.7\% to 80.5\%, without using any extra data, model, trick, or changing the structure, which surpasses existing individual training ingredients by a large margin. When adding only uniform augment~\cite{lingchen2020uniformaugment} to stimulative training++, the performance can be further improved to 81.0\%, better than the best training recipes provided by Timm library and PyTorch official version that combine plenty of ingredients with carefully-tuned hyper-parameters. Moreover, we not only verify the effectiveness of the stimulative training++ pretrained model on various downstream tasks, but also extend the stimulative training method to enhance the finetuning phase of downstream tasks.
We also present more experiments to verify the superiority of the proposed method, including employing kinds of models and datasets and testing the performance under different training costs.
In addition, we theoretically show the connection between the proposed method and the improved performance of a given residual network and its subnetworks. 

Compared to our previous NeurIPS conference version \cite{ye2022stimulative}, the improved version (termed as stimulative training++) further: (1) proposes a novel KL- loss that only aligns the logits direction between the main network and various subnetworks; (2) provides sampled subnetworks with random smaller inputs instead of the same input resolution of the main network; (3) explores the inter-stage sampling rules for better designing the sampling space; (4) conduct plenty of experiments and analysis to verify why and how the above three strategies work; (5) achieve new state-of-the-art results without bells and whistles, better than all individual training ingredients and all combined training recipes; (6) extends to various downstream tasks and verifies its effectiveness in both pretraining and finetuning phases; (7) gives more experiments, discussions, and findings.

The main contributions of this work can be summarized as the following:
\begin{itemize}
\vspace{-3pt}
\item We understand residual networks from a social psychology perspective, and find that different kinds of residual networks invariably suffer from the problem of network loafing.
\vspace{2pt}
\item We improve residual networks from a social psychology perspective, and propose a 
stimulative training scheme to improve the performance of a given residual network and its subnetworks.
\vspace{2pt}
\item We further design three simple-yet-effective strategies to unleash the potential of stimulative training, including novel KL- loss, random smaller inputs for subnetworks, and inter-stage sampling rules.
\vspace{2pt}
\item Comprehensive experiments and analysis verify the effectiveness of stimulative training as well as its three strategies. Besides, we show that the proposed method can not only achieve new SOTA results but also generalize well to various downstream tasks.
\vspace{2pt}
\item We present more experiments and findings as well as some theoretical analysis to show the rationality and superiority of the proposed training scheme.
\vspace{-3pt}
\end{itemize}

The remainder of this paper is organized as follows. Section 2 introduces the related works. Section 3 presents the proposed stimulative training and its three improved strategies. Section 4 shows the experimental verification. Section 5 gives the theoretical analysis. Section 6 concludes this paper. Section 7 points out possible limitations.

\section{Related Works} \label{sec:related work} 
\subsection{Unraveled View} \vspace{-1mm}
As one pioneer work to investigate residual networks,~\cite{veit2016residual} 
experimentally shows 
that residual networks can be seen as a collection of numerous networks of different lengths, namely unraveled view. Subsequently,~\cite{sun2022low} follows this view and further attributes the success of residual networks to shallow subnetworks first. In addition,~\cite{sun2022low} considers the neural network as a polynomial function and corresponds shallow subnetwork to low-degree term to explain the working mechanism of residual networks. Further,~\cite{barzilai2022kernel} theoretically proves that the eigenvalues of residual convolutional neural tangent kernel (CNTK) are composed of weighted sums of eigenvalues of CNTK of subnetworks. Similar to~\cite{veit2016residual},~\cite{sun2022low}, and~\cite{barzilai2022kernel}, this paper also investigates residual networks from the unraveled view. Differently, inspired by social psychology, we further reveal the loafing problem of residual networks under the unraveled view. Besides, we propose a novel stimulative training method to relieve the network loafing problem and improve the performance of residual networks. We further design three new strategies to unleash the potential of stimulative training.

\subsection{Knowledge Distillation}  \vspace{-1mm}
As a classical method, knowledge distillation transfers the knowledge from a teacher network to a student network via approximating the logits output~\cite{hinton2015distilling,qiu2022better,kim2018paraphrasing,mirzadeh2020improved} or/and features output~\cite{romero2014fitnets,huang2017like,heo2019knowledge, xu2018pad}. To avoid the huge cost of training a high-performance teacher, some works abandon the naive teacher-student framework, like mutual distillation~\cite{zhang2018deep} making groups of students learn from each other online, and self distillation~\cite{zhang2019your} transferring knowledge from deep layers to shallow layers. Generally, all these distillation methods need to introduce additional networks or structures, and employ fixed teacher-student pairs. In contrast, our method does not require any additional network or structure, and the student network is a randomly sampled subnetwork of a residual network. 
Besides, our method is essentially designed to address the loafing problem of residual networks, which is different from knowledge distillation that aims to obtain a compact network with acceptable accuracy.  
Technically, we further propose a novel KL- loss to only align the direction between the teacher logits and various students' logits, replacing commonly used KL loss that aligns both the direction and the magnitude of teacher-student network logits.
\subsection{One-shot NAS} \vspace{-1mm}
One-shot NAS is an important branch of neural architecture search (NAS)~\cite{ye2022b,chen2021bn,ye2022efficient,li2020improving,ye2022beta,guo2022neural} that aims to drastically reduce the training and search time for the whole pipeline. Along this direction,~\cite{cai2019once} trains a once-for-all (OFA) network with progressive shrinking and knowledge distillation to support all kinds of architectural settings. Following this work, BigNAS~\cite{yu2020bignas} introduces several technologies to train a high-quality single-stage model, whose child models can be directly deployed without extra retraining or post-processing steps. Both OFA and BigNAS aim at simultaneously training and searching various networks with different resolutions, depths, widths and operations. Differently, the proposed method aims at improving a given residual network and can be seamlessly applied to the searched model of NAS. In addition, as OFA and BigNAS are not designed to solve the loafing problem and improve the main network results, their subnet input designs, sampling rules, and supervision signals are all different from the proposed method.
\subsection{Stochastic Depth} \vspace{-1mm}
As a regularization technique, Stochastic Depth~\cite{huang2016deep} randomly disables the convolution layers of residual blocks, to reduce training time and test error substantially. In Stochastic Depth, the reduction in test error is attributed to strengthening gradients of earlier layers and the implicit ensemble of numerous subnetworks of different depths. In fact, its improved performance can be also interpreted as relieving the network loafing problem defined in this work. 
The theoretical analysis in our work can be also applied to explain Stochastic Depth. Besides, for better boosting the performance of residual network via training its subnetworks, 
we sample suitable subnetworks via well-designed inner-stage and inter-stage rules, provide random downscaling inputs for subnetworks, and use an additional KL- loss to yield a more achievable target which only makes the logits output direction of the residual network and its subnetworks more consistent.

\begin{figure*}[t]
\centering
\includegraphics[height=6.4cm]{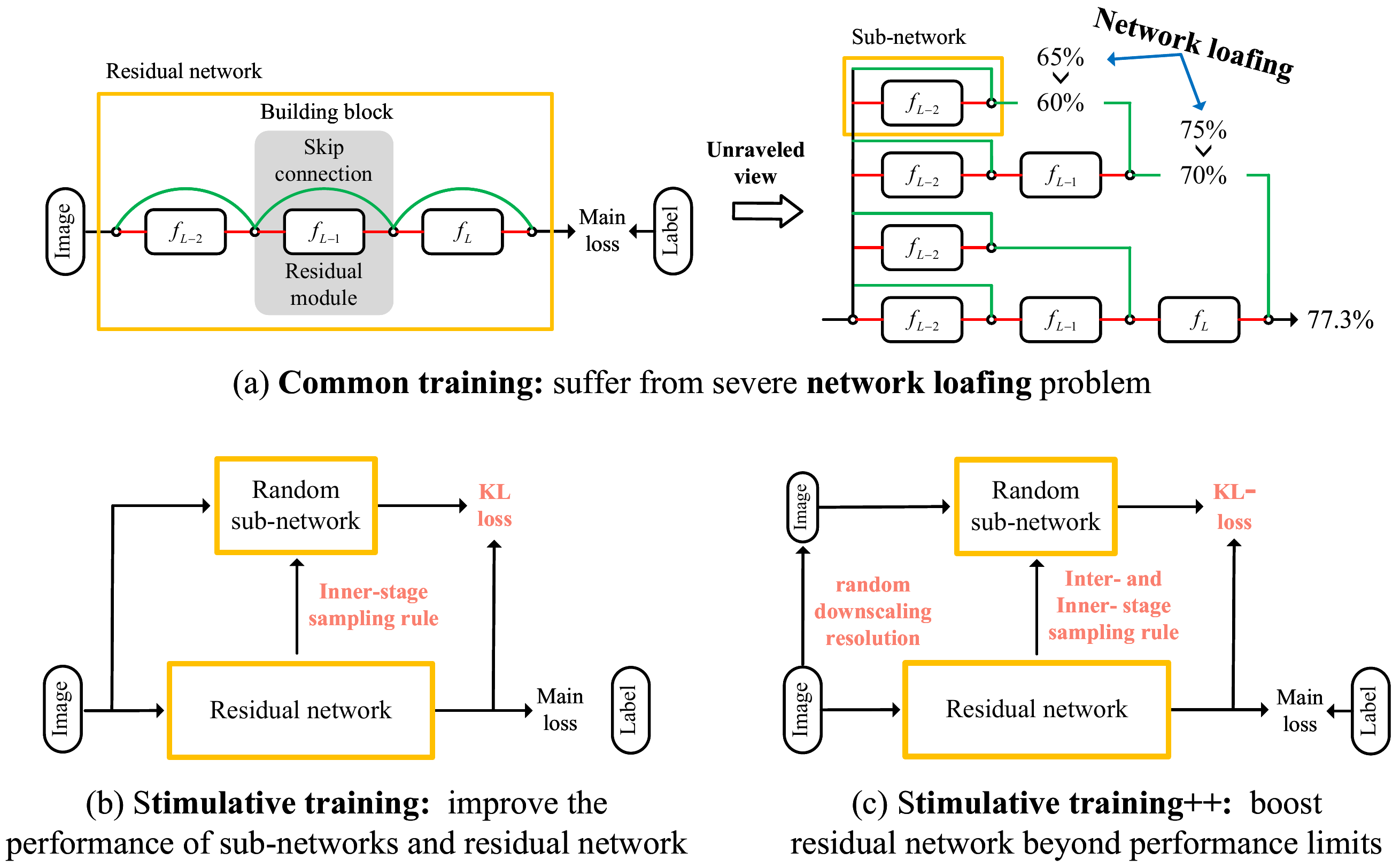}
\vspace{-2mm}
\caption{Illustration of common training, stimulative training~\cite{ye2022stimulative}, and stimulative training++ schemes. Common training suffers from severe network loafing problem. To relieve this, our conference version of  stimulative training~\cite{ye2022stimulative} enhances the performance of a residual network by improving its subnetworks. Stimulative training++ further proposes three strategies, boosting residual networks beyond their performance limits.
}
\label{fig:UT scheme}
\end{figure*}

\subsection{Training Ingredients and Recipes} \vspace{-1mm}
\label{section:Training Ingredients and Recipes}
Since the inception of AlexNet~\cite{krizhevsky2017imagenet}, training ingredients and recipes have made remarkable progress. Various training ingredients have been proposed, which can be roughly divided into following aspects: (1) Training schedules like longer epochs, larger batch sizes, and kinds of learning rate adjustment schemes~\cite{dosovitskiy2020image,tan2019efficientnet,touvron2021training}; (2) Data augmentation technologies like cutout~\cite{devries2017improved}, mixup~\cite{zhang2017mixup}, cutmix~\cite{yun2019cutmix}, rand erasing~\cite{zhong2020random}, rand augment~\cite{cubuk2020randaugment}, auto augment~\cite{cubuk2019autoaugment}, uniform augment~\cite{lingchen2020uniformaugment}, and trivial augment~\cite{muller2021trivialaugment}; (3) Network regularization technologies like StochDepth~\cite{huang2016deep}, Droppath~\cite{larsson2016fractalnet}, Dropblock~\cite{ghiasi2018dropblock}, and ShakeDrop~\cite{yamada2018shakedrop}; (4) New objectives like label smoothing~\cite{yuan2020revisiting}, and alternative loss functions of cross-entropy~\cite{beyer2020we,khosla2020supervised}; (5) New optimizers like SGD with Nesterov momentum~\cite{sutskever2013importance}, RMSProp~\cite{szegedy2016rethinking}, AdamW~\cite{loshchilov2017fixing}, and Lamb~\cite{you2019large}; (6) Other technologies like weight averaging~\cite{izmailov2018averaging}, repeated augmentation~\cite{berman2019multigrain}, and train-test resolution fixing~\cite{touvron2019fixing}. Each technique can obtain some improvements over the baseline (e.g., ResNet50). Recently, a few works attempt to provide the possible best training recipes by finding the best combination of these training ingredients, like bag of tricks~\cite{he2019bag}, timm library~\cite{wightman2021resnet}, and pytorch library~\cite{vryniotis2021train}. In this paper, we show that: (1) Our method can perform much better than any individual technique. (2) Our method can reach a new SOTA, with only uniform augment.

\section{The Proposed Method} \label{sec:method}
\subsection{Motivation} \vspace{-1mm}

Social loafing is a social psychology phenomenon that individuals lower their productivity when working in a group. Based on the novel view that a residual network behaves like an ensemble network~\cite{veit2016residual}, we find that various residual networks invariably exhibit loafing-like behaviors as shown in Fig.~\ref{fig:loafing_diff_networks}, which we define as \textbf{network loafing}.
As social loafing is ``a kind of social disease" that will harm the individual and collective productivity~\cite{latane1979many}, we consider network loafing may also hinder the performance of a residual network and its subnetworks. As shown in Fig.~\ref{fig:UT scheme} (a),
we find \textbf{common training} that only focuses on the optimization of main network suffers from severe network loafing. That is, subnetworks deteriorate their performance when working in an  ensemble-like residual network.
For example, subnetworks within the residual network only have an accuracy of 60\% and 70\%, much lower than their individual accuracy of 65\% and 75\%.

To alleviate network loafing, it is intuitive to learn from social psychology. There are two common methods for solving the social loafing problem in sociology, namely, establishing individual accountability (i.e., increasing individual supervision) and making tasks cooperative (i.e., setting up the overall goal)~\cite{williams1991social,george1995asymmetrical}. Inspired by this, we propose a novel \textbf{stimulative training} of residual networks in our conference version~\cite{ye2022stimulative}. To increase individual supervision, we randomly sample a subnetwork from the whole network at each iteration and provide extra supervision to train the subnetwork. For the overall goal, we adopt KL divergence loss to constrain the output of each sampled subnetwork not far from that of the whole network, making the output of all subnetworks and the whole network more consistent.

To further unleash the potential of stimulative training, we improve it from three different perspectives and propose three simple-yet-effective strategies, boosting residual networks beyond their performance limits, namely \textbf{stimulative training++}: (1) a novel KL- loss that only aligns the logits direction of the main network and various subnetworks is proposed to replace naive KL loss, based on the finding that different networks produce distinctive logits amplitude; (2) we provide the subnetworks with random smaller inputs instead of using the same input resolution with the main network, based on the observation that random smaller inputs can provide sampled subnetworks with multi-scale and larger effective receptive fields;
(3) we explore the important but previously ignored designing of inter-stage sampling rules of residual networks, based on the fact that inter-stage layers are not uniformly distributed 
in most networks.

\begin{figure*}[t]
\centering
\includegraphics[height=5cm]{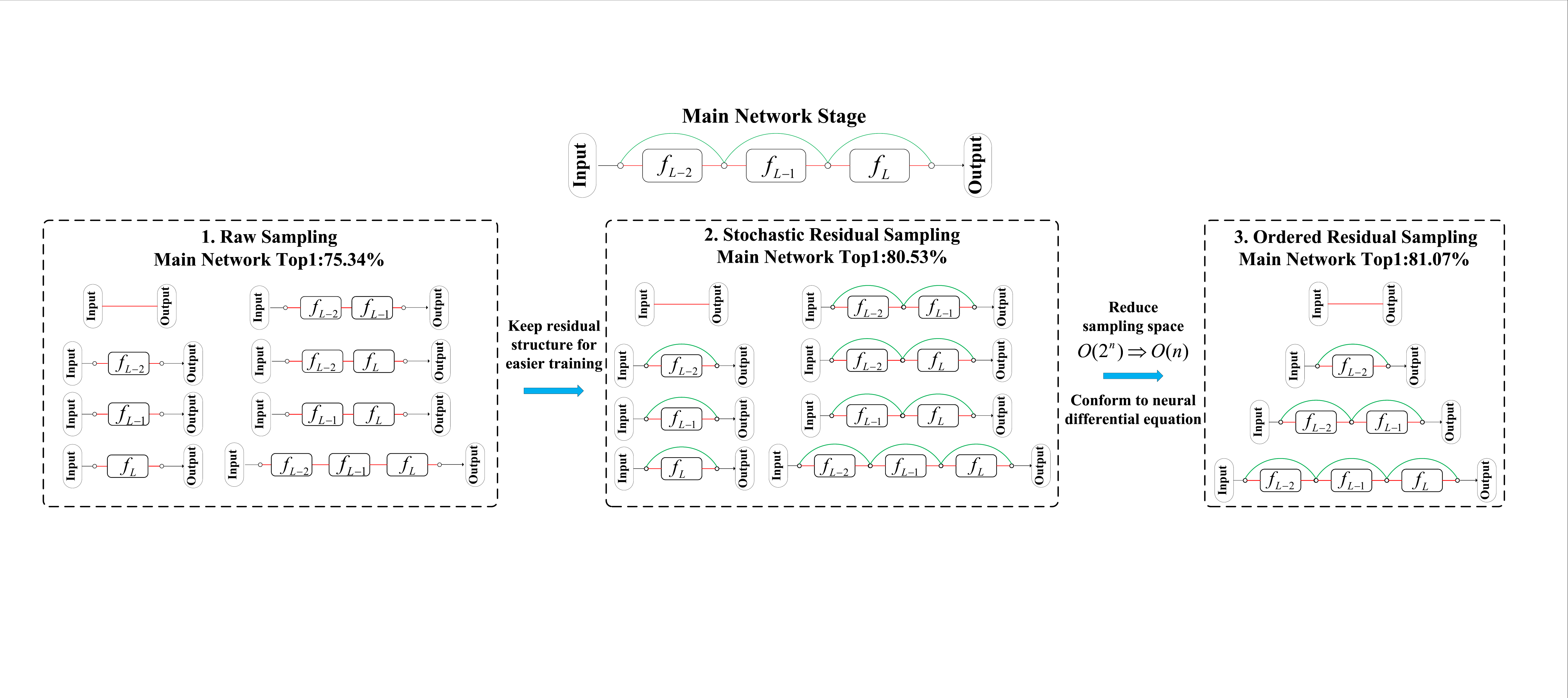}
\vspace{-3mm}
\caption{Illustration of the proposed ordered residual sampling. Compared with two other sampling strategies, ordered residual sampling can facilitate training and reduce the sampling space to get better main network performance. }
\label{fig:sampling way}
\vspace{-1mm}
\end{figure*}

\subsection{Stimulative Training} \vspace{-1mm} 
\subsubsection{Training Algorithm} \vspace{-1mm}
In this subsection, we briefly illustrate the pipeline of the proposed stimulative training, which optimizes the main network and meanwhile uses it to provide extra supervision for a sampled subnetwork at each training iteration, as shown in Fig.~\ref{fig:UT scheme} (b).
Formally, for a residual network to be optimized, we define the main network as $D_m$ and the subnetwork as $D_s$. All the subnetworks share weights with the main network, and make up the sampling space $\Theta = \left \{ D_s|D_s = \pi (D_m) \right \}$, where $\pi()$ is the random sampling operator.
To make the training more efficient and effective, we define a new sampling space obeying the ordered residual sampling rule to be discussed in Section~\ref{sec:ors}. Denoting $\theta_D{}_m$ and $\theta_D{}_s$ as the weights of the main network and sampled network respectively, $x$ as the mini-batch sample, and $y$ as its label, $\mathcal{Z}$ as the network output, the total loss of stimulative training for classification is computed as
\begin{equation}
\label{fomulation:ut loss} \mathcal{L}_{mt} = \underbrace{ CE(\mathcal{Z}(\theta_{D_m},x ),y)}_{\rm{main\ supervision}} + \underbrace{\lambda KL(\mathcal{Z}(\theta_{D_m},x),\mathcal{Z}(\theta_{D_s} ,x))}_{\rm {extra\ overall \ goal \ supervision}}
\end{equation}
where $CE$ and $KL$ denote the standard cross entropy loss and KL divergence loss respectively,  $\lambda$ is a balancing coefficient. $CE$ loss denotes the main supervision for the main network.
$KL$ loss provides extra supervision for the sampled subnetwork and guarantees all subnetworks to be optimized in the same direction as the main network, which can be considered as setting a common goal for the ``ensemble network”. 
We use standard stochastic gradient descent for updating the model, denoted as
$
\theta_{D_m}^{t+1} = \theta_{D_m}^{t} - \eta \frac{\partial \mathcal L_{ut}}{\partial \theta_{D_m}} 
$
, where $\eta$ is the learning rate. Thanks to the weight-sharing between subnetworks and the main network, we only need to update $\theta_{D_m}$ once for each iteration to optimize both the weights of $\theta_{D_m}$ and $\theta_{D_s}$. 

\subsubsection{Ordered Residual Sampling} \vspace{-1mm} \label{sec:ors}
Network loafing is based on the novel view that residual networks can be seen as the ensemble of numerous different subnetworks\cite{veit2016residual}. It is intuitive to adopt a raw unraveled view to design sampling space, namely raw sampling. 
However, there exist two problems in raw sampling. First, 
it introduces a mass of single branch subnetworks hard to optimize, as shown in the left sub-figure of Fig.~\ref{fig:sampling way}. 
Second, with the growth of network depth, the size of the sampling space increases exponentially. For a residual stage of $n$ blocks, there are $2^n$ subnetworks, which constitute a too large space to explore. Under limited computation resources, it is difficult to provide sufficient supervision to each subnetwork, causing insufficient stimulative training.

In order to facilitate stimulative training, we re-design two different sampling ways: stochastic residual sampling and ordered residual sampling. Fig.~\ref{fig:sampling way} shows a three-block residual network as a simple illustration for three sampling ways. For raw sampling, we obey the definition of~\cite{veit2016residual}, which means 
each subnetwork is composed of stacked convolution layers. For stochastic residual sampling, we keep the basic structure of residual blocks and skip some blocks randomly. For ordered residual sampling, we also keep the residual structure but skip blocks orderly, (i.e., always skip the last several blocks). We implement these sampling methods for MobileNetV3~\cite{howard2019searching} on CIFAR-100. Experimental results show that ordered residual sampling performs better than the other two methods (by 5.77\% and 0.58\%, respectively). To explain the superiority of ordered residual sampling, we should pay attention to keeping residual structure and ordered sampling. For keeping residual structure, it can drive the network training process easier and improve the final performance~\cite{he2016deep}.
For ordered sampling, it can noticeably reduce the size of sampling space (from $O(2^n)$ to $O(n)$), making it possible to train each subnetwork sufficiently. Besides, all deep networks with residual connections can be approximated by neural differential equations~\cite{chen2018neural}, which treat discrete network layers as continuous and regard the layer output as the result of adding an integral to the layer input. From this perspective, ordered sampling will maintain the continuity of residual networks. 
Because of these outstanding features, we will adopt ordered residual sampling in experiments by default.

\begin{figure*}[t]
  \centering
  \subcaptionbox{ResNet on CIFAR10}{
    \label{fig:amplitude_resnet_c10} 
    \includegraphics[width=0.32\linewidth]{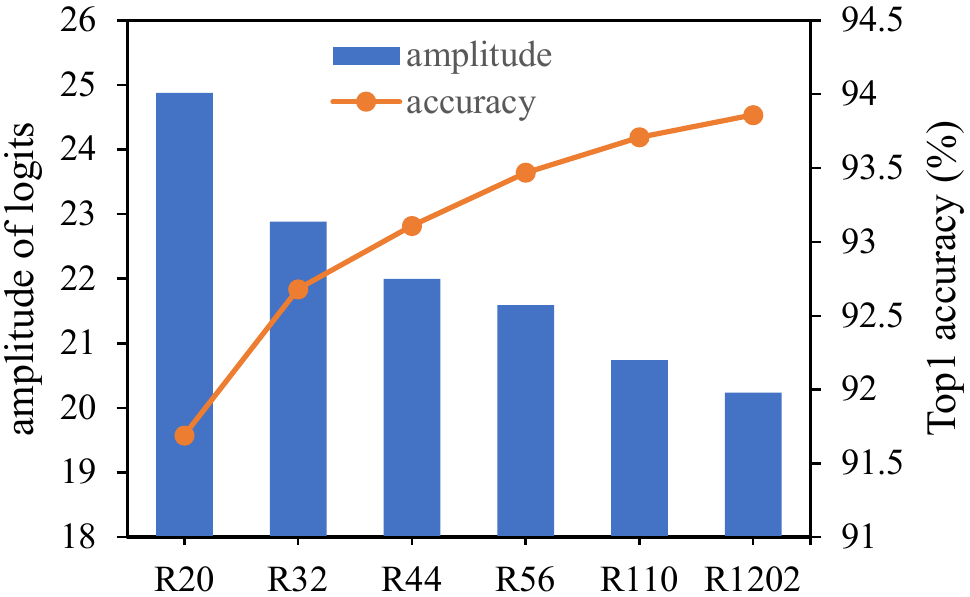}}
  \subcaptionbox{DenseNet on CIFAR100}{
    \label{fig:amplitude_densenet_c100} 
    \includegraphics[width=0.33\linewidth]{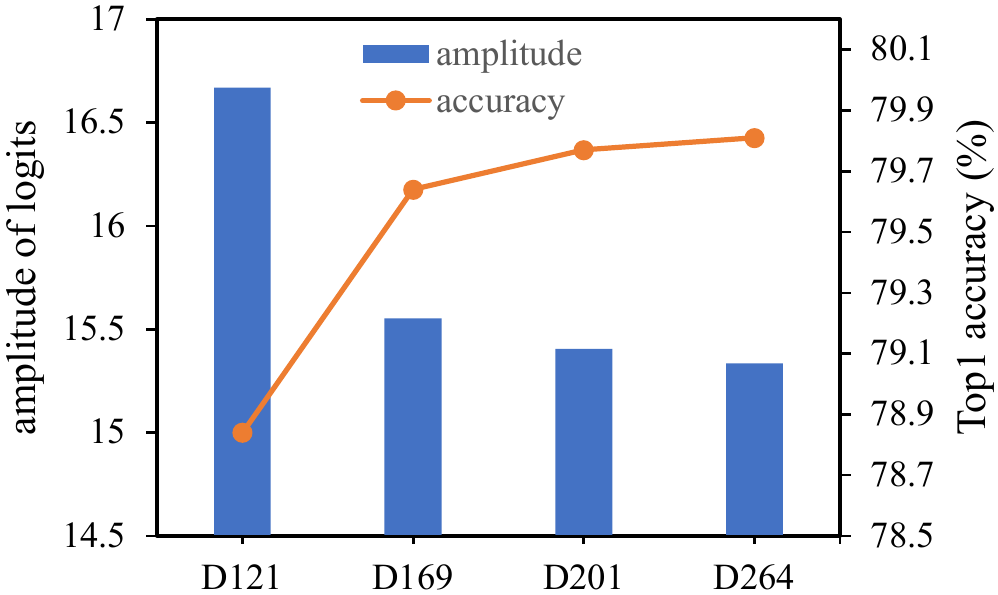}}
  \subcaptionbox{DenseNet on ImageNet}{
    \label{fig:amplitude_densenet_img} 
    \includegraphics[width=0.32\linewidth]{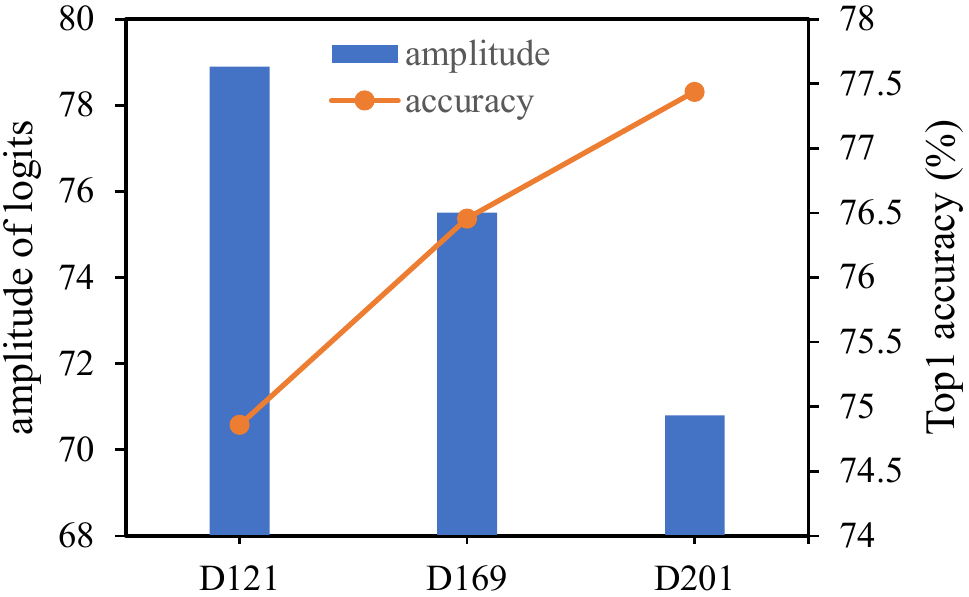}}
  \vspace{-2mm}
  \caption{The logits amplitude and Top1 accuracy  produced by different residual networks on different datasets are distinct after training. The horizontal axis means the different networks in a network family, such as ResNetx or DenseNetx.
  }
  \label{fig:amplitude_diff_networks} 
\vspace{-1mm}
\end{figure*}

\subsection{Stimulative Training++} 
\vspace{-1mm}
Stimulative training (ST) is an intuitive and primary method to alleviate network loafing via subnetworks training. To further unleash the potential of ST, we dive into the characteristics of subnetworks under unraveled view and observe three important phenomena: (1) networks with different capacities tend to produce distinctive amplitude of logits; 
(2) smaller networks usually have more narrow effective receptive fields (ERF); (3) inter-stage layers in most networks are not uniformly distributed while sampling networks plays an important factor in ST. We consider these phenomena may affect the effectiveness of ST and constrain the performance of residual networks. 
Thus, we further propose stimulative training++ (ST++), which improves ST with three simple-but-effective strategies: (1) knowledge distillation without magnitude; (2) subnetworks with random smaller Inputs; (3) inter-stage sampling rule. With these strategies, ST++ can boost residual networks beyond their performance limits without bells and whistles, better than those trained with any training ingredient or training recipe.
We will detailedly introduce these strategies in the remainder of this section.

\subsubsection{Knowledge Distillation Without Magnitude}
Recently, Wei et. al~\cite{wei2022mitigating} decompose the network logits into logit direction and logit magnitude, and find that network logit magnitude does not change the predicted class while highly affecting the network confidence degree. 
Inspired by this, we independently train a number of residual networks from scratch. After training, we accumulate logit amplitudes of each network in the whole dataset. As shown in Fig.~\ref{fig:amplitude_diff_networks}, even in the same network families, the logit amplitudes of networks with different capacities and Top1 accuracy are distinct, and such phenomena exist across different networks and datasets. Between different network families, the difference in logit amplitudes is more obvious. Based on the above observations, we conclude that different networks may have different logit amplitudes (confidence degrees). 

In original ST, the KL divergence loss is adopted to align the output probability of the main network and sampled subnetwork. In other words, original ST aligns both the logit direction and logit magnitude between the main network and sampled subnetwork. According to the above findings, it may be suboptimal 
to align the logit magnitude between diverse subnetworks and the same main network. Such a problem will become more serious in ST, due to the shared classifier of the main network and all subnetworks. Since aligning the logit direction (predicted class) between teacher and student is what we really care about, we propose to eliminate the influence of logit magnitude when computing the KL divergence loss between the main network and sampled subnetwork in ST.
Such a process can be regarded as knowledge distillation without magnitude, and the optimization target can be seen as KL without magnitude (KL-) loss. To illustrate the novel KL- loss more clearly, we formalize KL- as follows. 

\begin{figure}[t]
\centering
\includegraphics[width=0.4\textwidth]{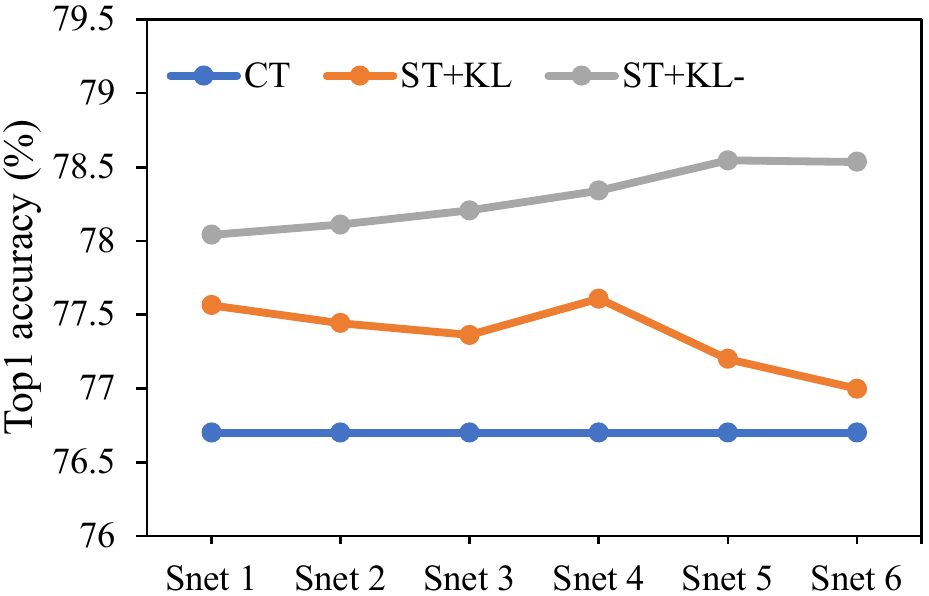}
\vspace{-2mm}
\caption{Verification of two superior properties of KL without magnitude (KL-). ST and CT refer to stimulative training and common training. From the table, we observe: (1) ST with KL- can significantly improve the result compared to ST with KL; (2) As the number of sampling subnetworks increases at each iteration, ST with KL- consistently increases the performance, while ST with KL may degrade the performance.
}
\label{fig:kd_vs_kd-}
\vspace{-1mm}
\end{figure}

\begin{table*}
\centering
\caption{Performance comparison under different kinds of subnet input transforms (SITrans). With smaller inputs for subnets, the performance consistently increases. With larger inputs for subnets, the performance consistently decreases. With other transforms for subnets, the performance may remain the same or decrease.}
\label{table:SITrans}
\vspace{-6pt}
	\begin{minipage}{0.33\linewidth}
	\centering
        \subcaption{Smaller Inputs for Subnets}
        \label{subtable:smaller_inputs}
        \vspace{-3pt}
\resizebox{\textwidth}{16mm}{
\begin{tabular}{c|c|c}
\hline
Subnet Input Transform                                                                                      & Top-1 (\%) & Top-5 (\%) \\ \hline
None                                                                                                        & 77.564     & 93.852     \\ \hline
\begin{tabular}[c]{@{}c@{}}{[}Resize(random.choice(np.lins\\ pace(128, 224, 4))), normalize{]}\end{tabular} & 78.224     & 94.180     \\ \hline
\begin{tabular}[c]{@{}c@{}}{[}Resize(random.randint(128, \\ 224)), normalize{]}\end{tabular}                & 78.324     & 94.222     \\ \hline
\begin{tabular}[c]{@{}c@{}}{[}Resize(random.randint(96, \\ 224)), normalize{]}\end{tabular}                 & 78.562     & 94.222     \\ \hline
\begin{tabular}[c]{@{}c@{}}{[}Resize(random.randint(64, \\ 224)), normalize{]}\end{tabular}                 & 78.574     & 94.378     \\ \hline
\end{tabular}}
	\end{minipage}
	\hfill
	\begin{minipage}{0.33\linewidth}
	\centering
        \subcaption{Larger Inputs for Subnets}
        \label{subtable:larger_inputs}
        \vspace{-3pt}
\resizebox{\textwidth}{16mm}{
\begin{tabular}{c|c|c}
\hline
Subnet Input Transform                                                                     & Top-1 (\%) & Top-5 (\%) \\ \hline
None                                                                                       & 77.564     & 93.852     \\ \hline
\begin{tabular}[c]{@{}c@{}}{[}Resize(random.randint(96, \\ 224)),normalize{]}\end{tabular} & 78.562     & 94.222     \\ \hline
\begin{tabular}[c]{@{}c@{}}{[}Resize(random.randint(96, \\ 240)),normalize{]}\end{tabular} & 78.430     & 94.270     \\ \hline
\begin{tabular}[c]{@{}c@{}}{[}Resize(random.randint(96, \\ 256)),normalize{]}\end{tabular} & 78.350     & 94.196     \\ \hline
\begin{tabular}[c]{@{}c@{}}{[}Resize(random.randint(96, \\ 272)),normalize{]}\end{tabular} & 78.058     & 94.002     \\ \hline
\end{tabular}}
\end{minipage}
 	\hfill
  	\begin{minipage}{0.33\linewidth}
	\centering
        \subcaption{Other Transforms for Subnets}
        \label{subtable:other_trans}
        \vspace{-3pt}
\resizebox{\textwidth}{16mm}{
\begin{tabular}{c|c|c}
\hline
Subnet Input Transform                                                                                                                  & Top-1 (\%)              & Top-5 (\%)              \\ \hline
None                                                                                                                                    & 77.564                  & 93.852                  \\ \hline
{[}RandomResizedCrop(224){]}                                                                                                            & 77.536                  & 93.820                  \\ \hline
\begin{tabular}[c]{@{}c@{}}{[}RandomResizedCrop(224),\\ normalize{]}\end{tabular}                                                       & 77.648                  & 93.612                  \\ \hline
\begin{tabular}[c]{@{}c@{}}{[}RandomResizedCrop(224),\\ jittering, normalize{]})\end{tabular}                                           & 76.752                  & 93.170                  \\ \hline
\multirow{3}{*}{\begin{tabular}[c]{@{}c@{}}{[}RandomResizedCrop(224),\\ RandomHorizontalFlip(),\\ jittering, normalize{]}\end{tabular}} & \multirow{3}{*}{76.566} & \multirow{3}{*}{93.224} \\                  &                         &                         \\               &                         &                         \\ \hline
\end{tabular}}
\end{minipage}
\hfill
\end{table*}

Formally, original KL divergence loss can be written as
\begin{align}
\label{formulation:kl1}
\mathcal{L}_{KL} = KL(p^t,&p^s) =  \sum_{i=1}^{N} p_{i}^t \log \frac{p_{i}^t}{p_{i}^s} \\
\label{formulation:kl2}
p_i &= \frac {e^{Z_i}} {\sum_{j=1}^{N} e^{Z_j}} 
\end{align}
where $Z$ is the network output logits, $p$ is the classification probability after softmax, $N$ is the number of classes, $t$ and $s$ denote teacher and student respectively. 
Without loss of generality, $Z$ can be decomposed into logit magnitude and logit direction, which can be written as 
\begin{align}
 Z = ||Z|| \cdot \hat{Z}
\end{align} 
where $||Z||$ denotes the magnitude ($L_2$ norm) of $Z$, which can be computed as $||Z|| = \sqrt{\sum_{i=1}^N {Z_i}^2} $, and $\hat{Z} = \frac{Z}{||Z||}$ is the direction (unit vector). In this paper, in order to remove the possible negative effect of logit magnitude, we only utilize $\hat{Z}$ to compute the softmax-activated probability, and then compute the KL- loss as
\begin{align}
\mathcal{L}_{KL-} &= KL(\hat{p}^t,\hat{p}^s) =  \sum_{i=1}^{N} \hat{p}_{i}^t \log \frac{\hat{p}_{i}^t}{\hat{p}_{i}^s} \\
\hat{p}_i &= \frac {e^{\hat{Z_i}}} {\sum_{j=1}^{N} e^{\hat{Z}_j}}= \frac {e^{\frac{Z_i}{||Z||}}} {\sum_{j=1}^{N} e^{\frac{Z_j}{||Z||}}} 
\end{align} 
where $\hat{p}$ is the normalized classification probability after softmax. Compared with the original KL divergence loss, the proposed KL- loss only aligns the logit direction between the teacher and student and allows each network to have its own logit magnitude, which is more suitable for the proposed stimulative training that contains diverse students and shares the classifier.

\par To verify the superiority of KL-, we conduct ST with naive KL loss and novel KL- loss for ResNet50 network on the ImageNet dataset, respectively. We further compare their performances when sampling different numbers of subnetworks at each training iteration. The results are shown in Fig.~\ref{fig:kd_vs_kd-}. It is obvious that ST with KL- loss can consistently surpass ST with KL loss when sampling different numbers of subnetworks, which demonstrates the outstanding performance of KL-. More importantly, when sampling multiple subnetworks at each iteration, ST with KL generally degrades the performance. As a comparison, ST with KL- consistently increases the performance as the number of sampling subnetworks increases.

\vspace{-1mm}
\subsubsection{Subnetworks with Random Smaller Inputs}
Effective receptive field (ERF)~\cite{luo2016understanding} is a crucial factor affecting the performance of convolutional networks. A small ERF remarkably hinders the process of capturing the textural and structural information, as only limited regions in images can respond to the output of convolutional kernel~\cite{luo2016understanding}. As a generally accepted criterion in designing modern networks~\cite{he2016deep, howard2017mobilenets, huang2017densely, zhang2018shufflenet}, an effective method for building up large ERF is to stack more layers and make networks deeper. However, under the unraveled view of residual networks, there are numerous shallow subnetworks in the implicit ensemble system, leading to a concern that shallow subnetworks suffer from small ERF more seriously, which may limit the performance of subnetworks and ST. 

\begin{figure}[t]
  \centering
  \subcaptionbox{R50+224}{
    \label{fig:resnet50_heatmap_224} 
    \includegraphics[width=0.305\linewidth]{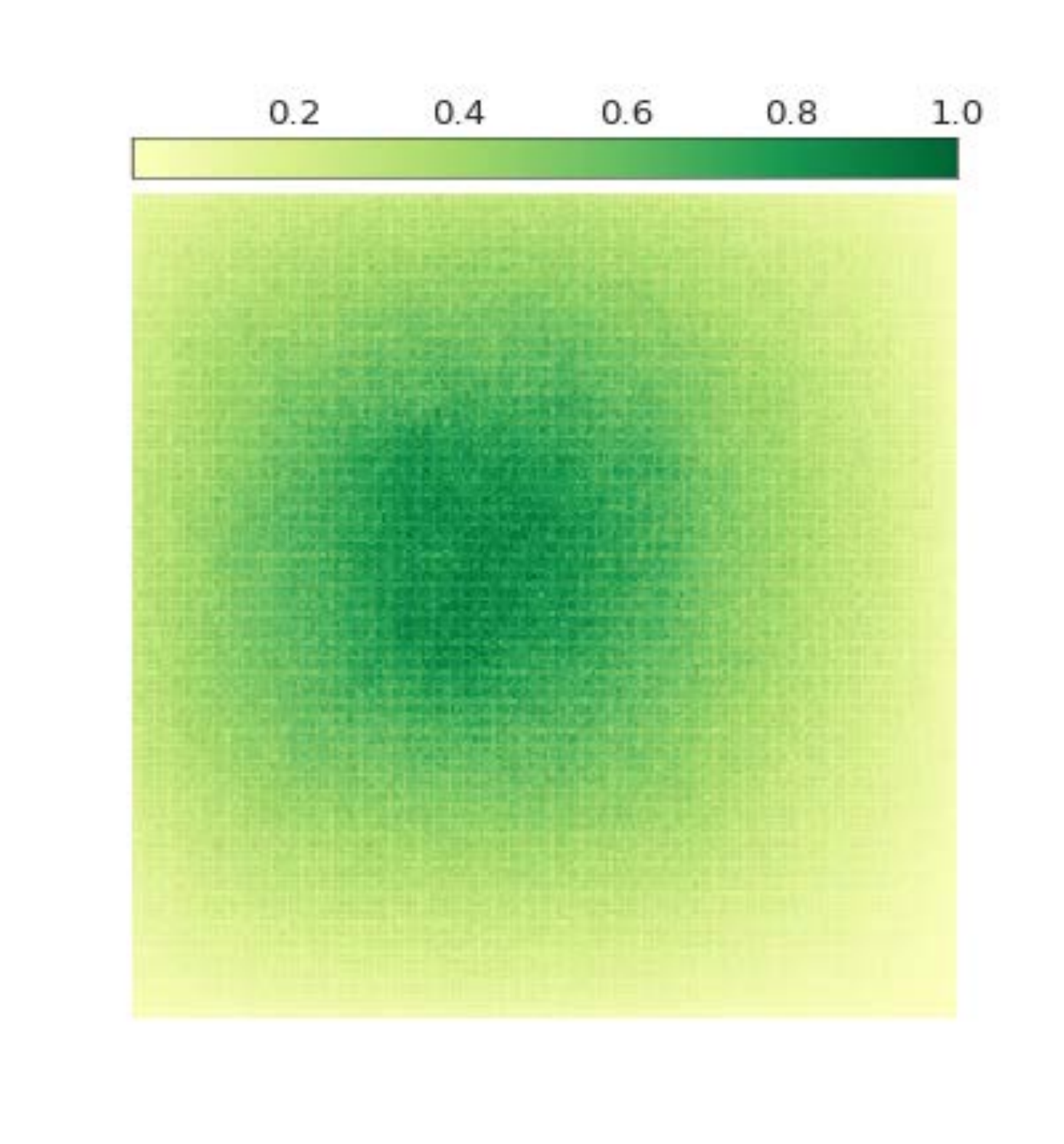}}
  \subcaptionbox{R101+224}{
    \label{fig:resnet101_heatmap_224} 
    \includegraphics[width=0.3\linewidth]{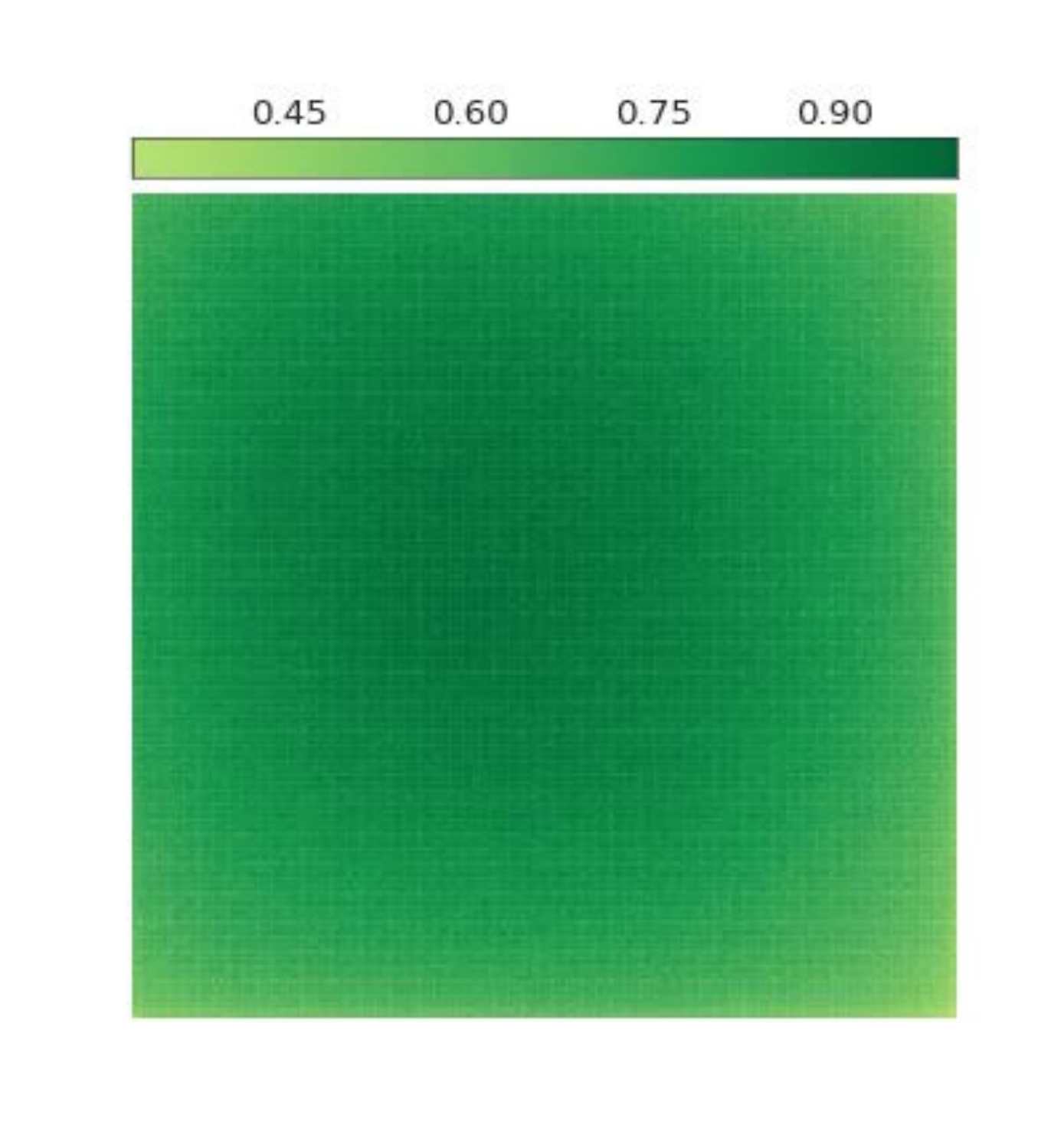}}
  \subcaptionbox{R152+224}{
    \label{fig:resnet152_heatmap_224} 
    \includegraphics[width=0.3\linewidth]{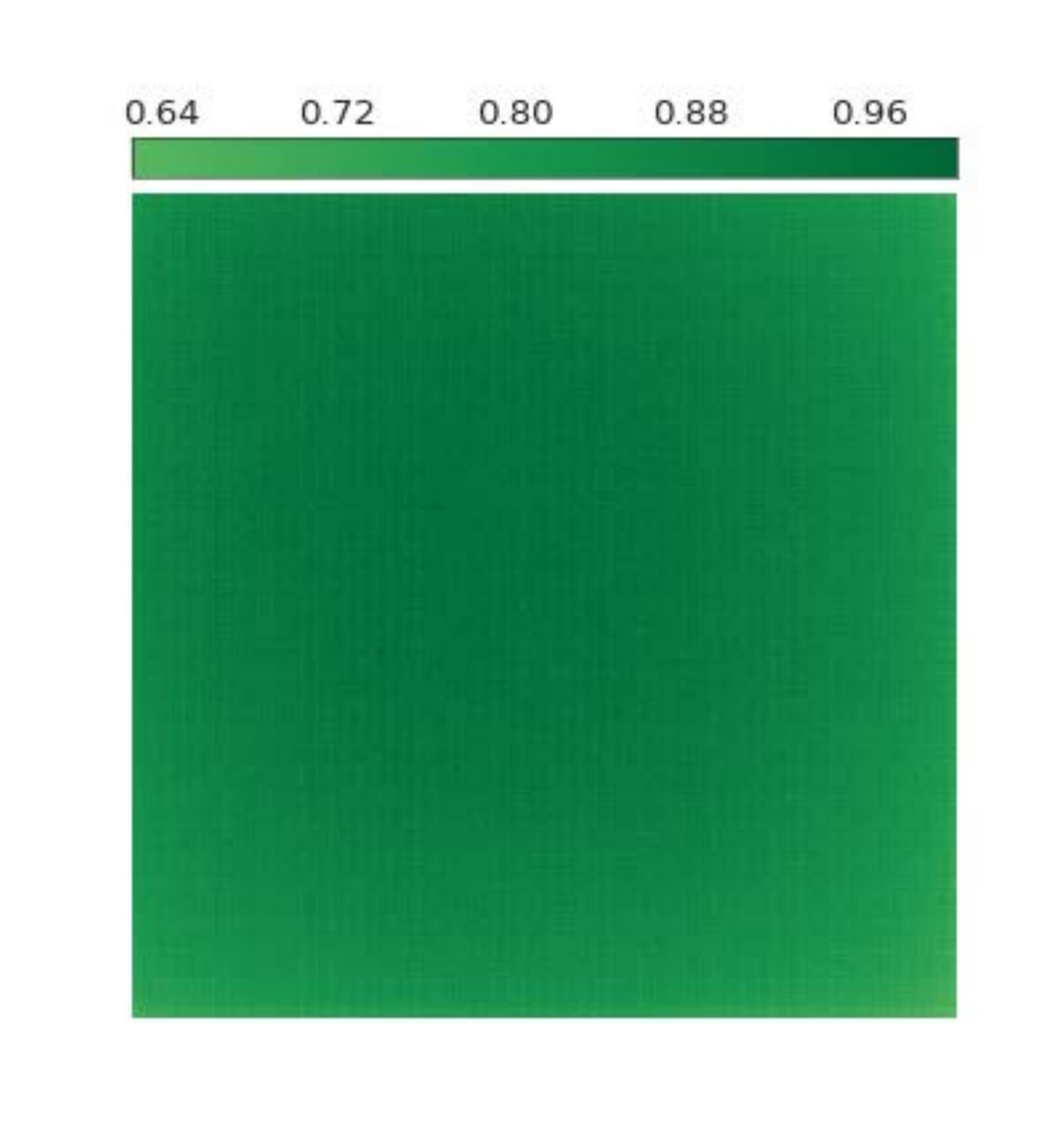}}
  \subcaptionbox{R50+512}{
    \label{fig:resnet50_heatmap_512} 
    \includegraphics[width=0.31\linewidth]{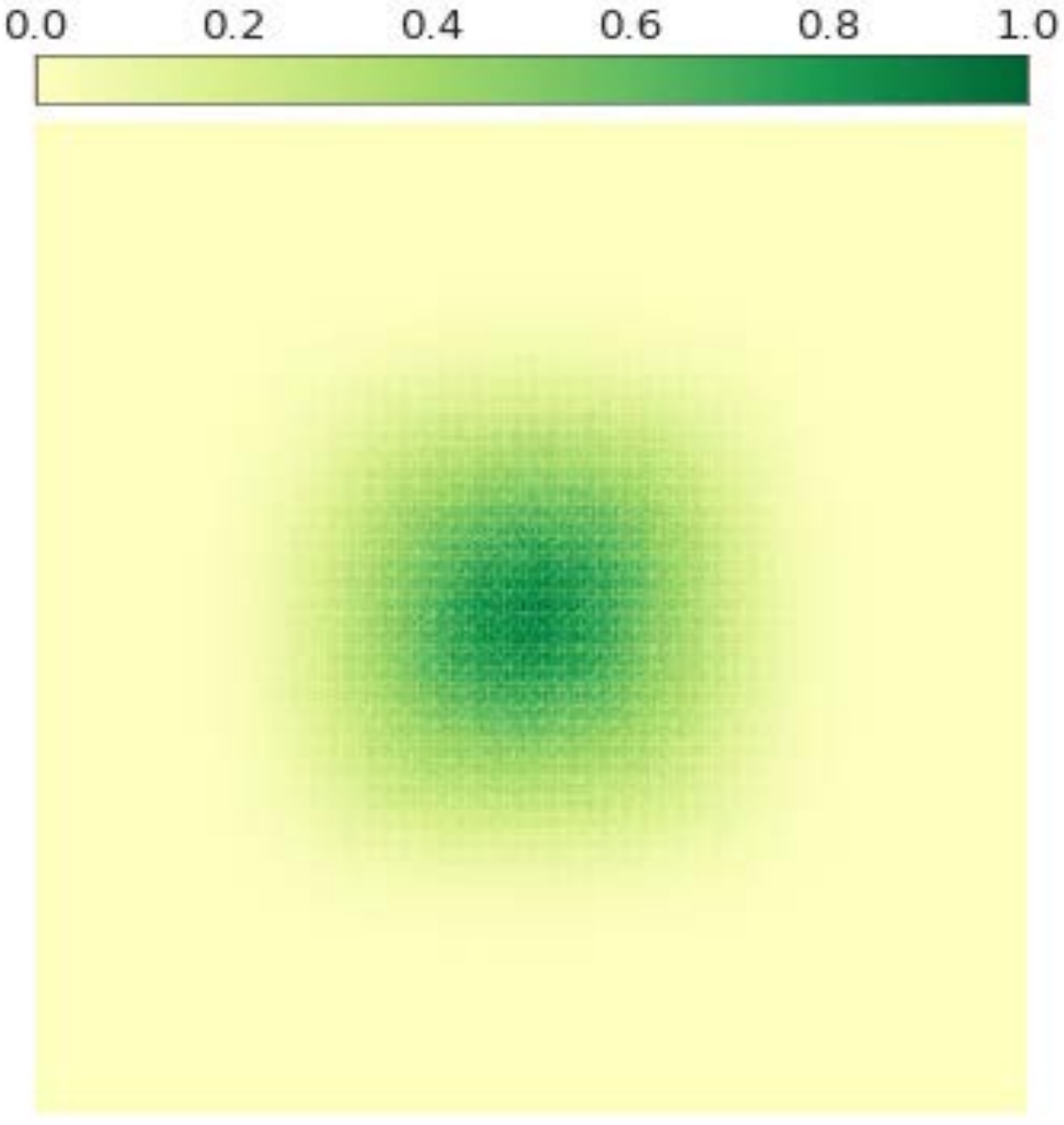}}
  \subcaptionbox{R101+512}{
    \label{fig:resnet101_heatmap_512} 
    \includegraphics[width=0.305\linewidth]{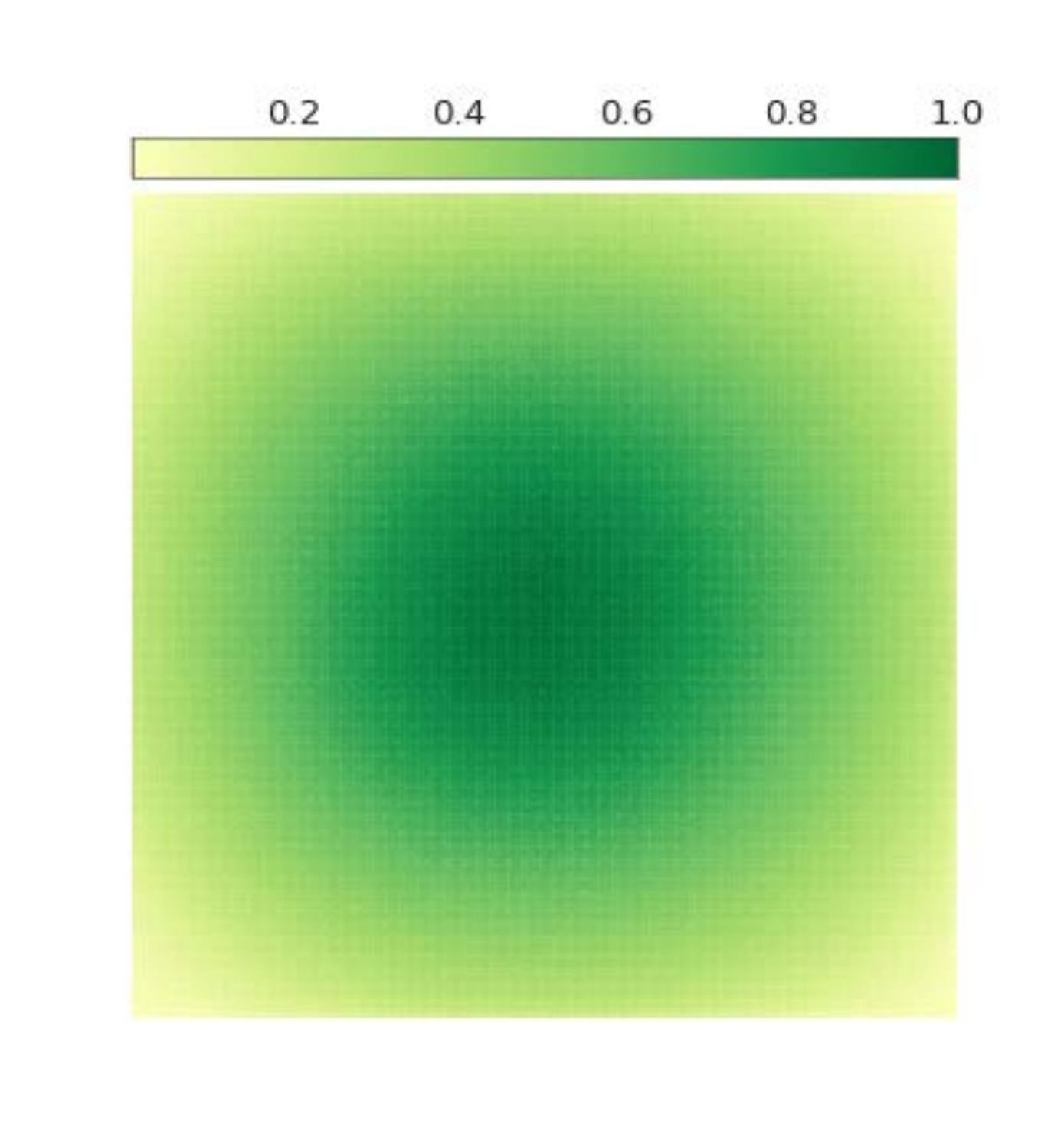}}
  \subcaptionbox{R152+512}{
    \label{fig:resnet152_heatmap_512} 
    \includegraphics[width=0.3\linewidth]{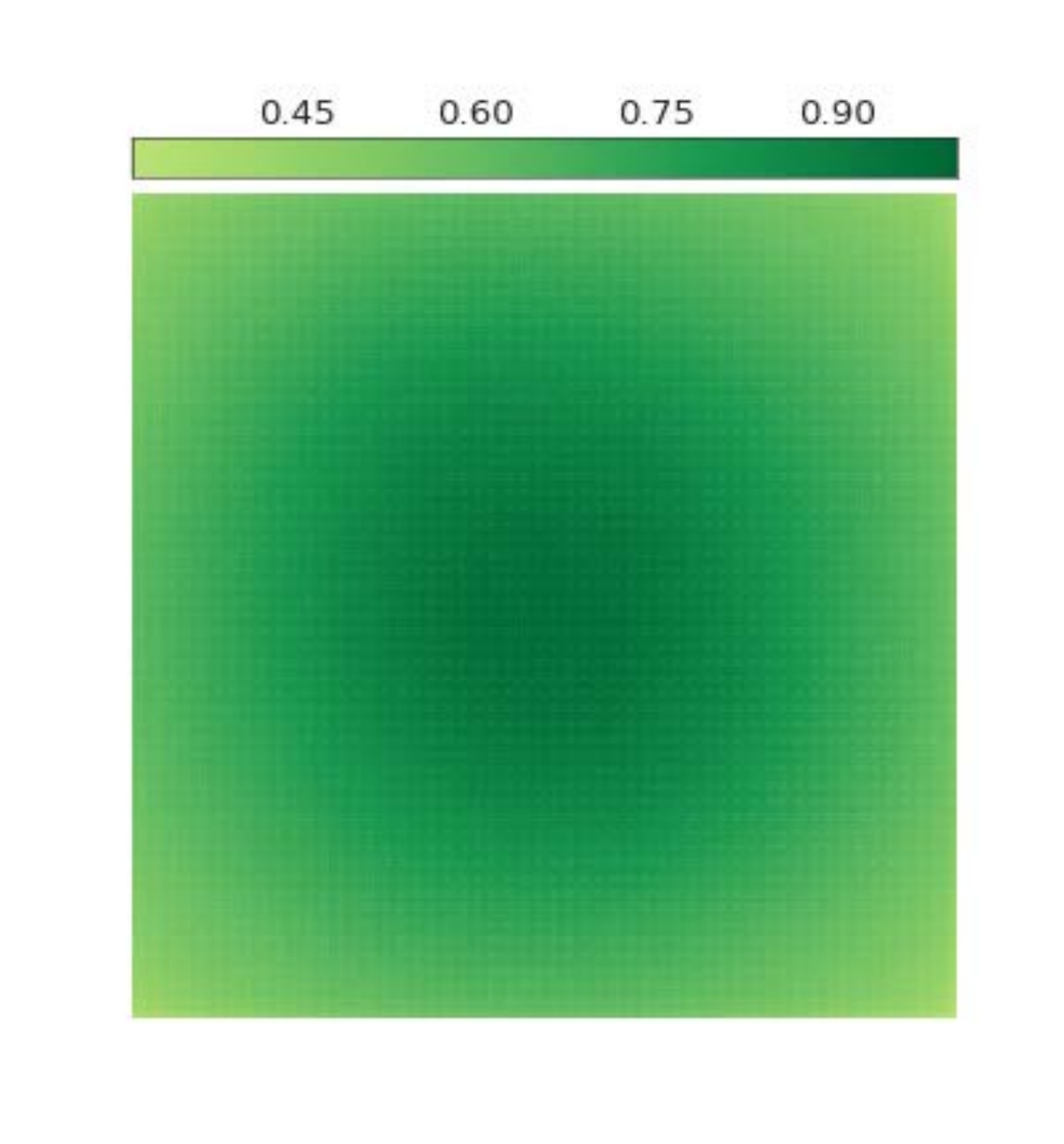}}
  \vspace{-2mm}
  \caption{Effective receptive fields (ERF) generated by different networks with different input resolutions. ERF is similar between large networks with high input resolutions and small networks with low input resolutions.
  }
  \label{fig:ERF_diff_networks} 
\vspace{-1mm}
\end{figure}

To clarify this, we follow the visualization method used in~\cite{ding2022scaling} and further study the ERF generated by different networks with different input resolutions. The results are shown in Fig.~\ref{fig:ERF_diff_networks}. The larger the ERF, the larger the dark area on the heatmap. 
It can be seen that deeper networks tend to generate larger ERF, which is in accord with the discovery in~\cite{ding2022scaling}. More interestingly, we observe that: (1) simply downscaling the input resolutions can remarkably enlarge the ERF of relatively shallow networks, and shallow networks with lower input resolutions can generate a similar ERF to deep networks with higher input resolutions; (2) when given inputs of different resolutions, the network can generate ERFs of different scales, which may help capture multi-scale information. Specifically, in Fig.~\ref{fig:ERF_diff_networks}, ResNet50 (or ResNet101) with $224\times224$ input resolution can generate a similar ERF to ResNet101 (or ResNet152) with $512\times512$ input resolution, and when inputting $224\times224$ and $512\times512$ resolutions to each ResNet, different scale ERFs are generated.

\begin{figure*}[t]
\centering
\includegraphics[width=0.98\textwidth]{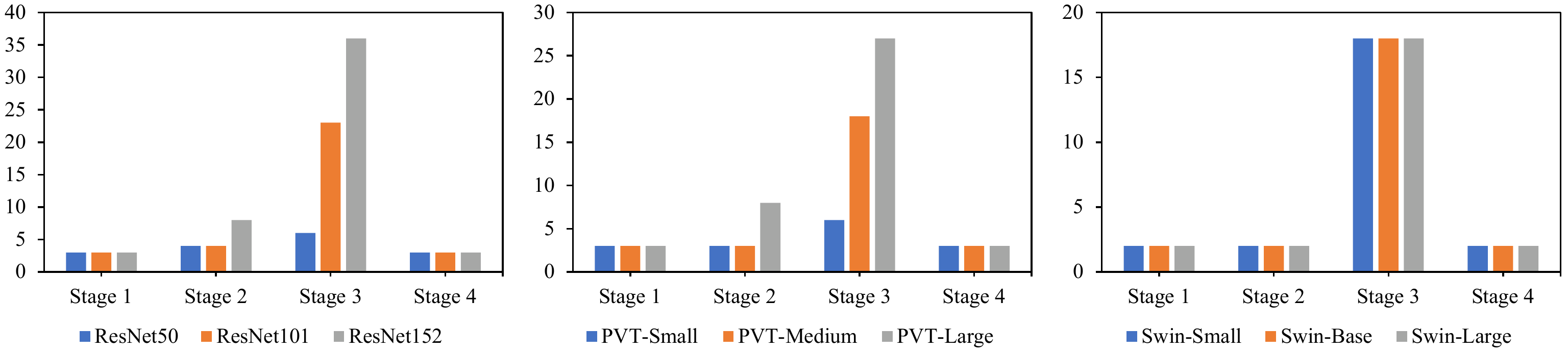}
\vspace{-3mm}
\caption{The layer number at different stages of different network families, such as ResNetx, PVT-x, and Swin-x. Generally, inter-stage layers are not uniformly distributed, and some stages (e.g., stage 2 and 3) tend to have much more layers.}
\label{fig:layer number}
\vspace{-1mm}
\end{figure*}

Inspired by the above observations, we propose to 
apply random downscaling inputs to subnetworks while keeping the input resolution of the main network the largest. The downscaling operation can enlarge the ERF of subnetworks, the random operation can provide subnetworks with multi-scale ERFs, they work together to improve the performance of subnetworks and ST. We name this strategy Subnetworks with Random Smaller Inputs and formalize it as follows. Without loss of generality, we define the shorter side of the image height and width as $L$. And the shorter side of input images for the main network and sampled subnetwork are denoted as $L_m$ and $L_s$, respectively. Let $U()$ represent the uniform random sampling, and $[L_{min},...,L_{max}]$ represent the sampling space consisting of integers, where $L_{max}$ and $L_{min}$ denote the largest and smallest integer respectively. 
Formally, we randomly sample an integer from $[L_{min},...,L_{max}]$ as $L_s$, and keep $L_m$ always equal to the shorter side of input image $x$, written as
\begin{align}
    L_s = U &([L_{min},...,L_{max}])
\end{align}
Actually, the training of the main network remains the same as before. Differently, we resize the original input image $x$ by standard bilinear interpolation before inputting it into the sampled subnetwork for optimization, written as
\begin{align}
    Z_s &= \mathcal{Z}(\theta_{D_s}, Resize(x;L_s)) \\
    Z_m &= \mathcal{Z}(\theta_{D_m}, x),
\end{align}
where $Resize(x;L_s)$ means to resize the shorter side of input image $x$ to $L_s$ while keeping the aspect ratio unchanged.

\par To comprehensively analyze the effect of different kinds of subnet input transforms (SITrans), we conduct original ST with: 1) Smaller Inputs for Subnets; 2) Larger Inputs for Subnets; 3) Other Transforms for Subnets.
All experiments are performed with ResNet50 on ImageNet, and the results are reported in Table~\ref{table:SITrans}. As shown in Table~\ref{subtable:smaller_inputs}, the performance of ST can be improved by either including more input resolution choices or decreasing the minimum input resolution for subnets, which validates our analysis that utilizing multi-scale ERFs or enlarging the ERF for subnets can benefit ST. As shown in Table~\ref{subtable:larger_inputs}, when we gradually increase the maximum input resolution of subnets, the performance of ST gradually decreases, which further verifies the importance of enlarging the ERF of subnets. To demonstrate the unique effects of using random downscaling inputs for subnets, we also apply other types of SITrans besides resizing for comparison. As shown in Table~\ref{subtable:other_trans}, with other SItrans for subnets such as color jittering and RandomHorizontalFlip, the performance of ST may remain the same or even decrease. It suggests that the performance improvement of using smaller input for subnets is not due to more intense data augmentation but utilizing multi-scale ERFs and enlarging the ERF for subnets.

\vspace{-1mm}
\subsubsection{Inter-stage Sampling Rule}
Although sampling subnetworks plays an important factor in ST, there is still a lack of guidelines for designing the sampling space. In the original ST, we propose to obtain random  subnetworks from the main network via ordered residual sampling, as shown in Fig.~\ref{fig:sampling way}.
While such ordered residual sampling can benefit ST, it may ignore the inter-stage layer distribution of residual networks and sample subnetworks without sufficient capacity.
To show this, we further report the layer number of different stages of popular residual network families, including ResNetx~\cite{he2016deep}, PVT-x~\cite{wang2021pyramid}, and Swin-x\cite{liu2021swin} in Fig.~\ref{fig:layer number}. 
As we can see, both Transformer-based and CNN-based networks commonly adopt a bottleneck-like design for the inter-stage layer, which is characterized by having more layers in the middle stage and fewer layers at both ends of the network. On the other hand, in order to handle some complex datasets such as ImageNet, these networks generally have more than one layer at each stage to ensure their basic capacity.

\begin{table}[]
\caption{Comparison of different inter-stage sampling rules (InterSSR). The results achieve the best when keeping more topology consistency between the main network and subnetwork and guaranteeing more basic capacity of each stage.
}
\vspace{-2mm}
\centering
\label{table:InterSSR}
\begin{tabular}{cc|c|c}
\hline
\multicolumn{2}{l|}{\multirow{2}{*}{Inter-stage sampling rule (InterSSR)}} & \multirow{2}{*}{Top1(\%)} & \multirow{2}{*}{Top5(\%)} \\
\multicolumn{2}{l|}{}                                           &                           &                           \\ \hline
\multicolumn{1}{c|}{sampling space}           & {[}3, 4, 6, 3{]}         & 80.038                    & 95.076                    \\
\multicolumn{1}{c|}{sampling space1}          & {[}2, 3, 6, 3{]}         & 80.198                    & 95.080                    \\
\multicolumn{1}{c|}{sampling space2}          & {[}2, 3, 5, 2{]}         & 80.350                    & 95.144                    \\
\multicolumn{1}{c|}{sampling space3}          & {[}1, 2, 4, 1{]}         & 80.496                    & 95.264                    \\
\multicolumn{1}{c|}{sampling space4}          & {[}1, 2, 5, 2{]}         & 80.380                    & 95.152                    \\
\multicolumn{1}{c|}{sampling space5}          & {[}1, 2, 4, 2{]}         & 80.334                    & 95.204                    \\ \hline
\end{tabular}
\vspace{-1mm}
\end{table}

\par Inspired by these observations, besides ordered residual sampling, we further propose an inter-stage sampling rule (InterSSR) as another guideline to design the sampling space of ST. 
More specifically, we prefer to sample stages that have more layers, while also retaining the first several layers of each stage. This approach helps to maintain topology consistency between the main network and the sampled subnetwork, while also ensuring that each stage has a sufficient level of basic capacity.
\par We experimentally verify the effectiveness of such inter-stage sampling rule in Table~\ref{table:InterSSR}. Compared with the original sampling space (i.e., first row), ensuring more basic capacity at different stages can obtain better results consistently, and the performance improvement on the ImageNet dataset ranges from 0.15\% with $[2, 3, 6, 3]$ to 0.46\% with $[1, 2, 4, 1]$. Further, we can see that the best results are obtained by both considering keeping more topology consistency between the main network and
sampled subnetwork and ensuring more basic capacity of the sampled subnetwork.
\vspace{-1mm}

\begin{table*}[]
\caption{Effectiveness of the proposed three strategies (i.e., KL-+Snet6, SITrans, and InterSSR). Note that validation size remains 224$\times$224 in all experiments. Vrs240 means validation resize size is 240$\times$240 (default: 256$\times$256), and Tcs200 means train crop size is 200$\times$200 (default: 224$\times$224). UA denotes the data augmentation strategy called uniform augment~\cite{lingchen2020uniformaugment}.}
\vspace{-2mm}
\centering
\label{table:Effectiveness of three strategies}
\begin{tabular}{l|c|c|c|c}
\hline
\multicolumn{1}{c|}{model}                          & Top-1 (\%) & $\Delta$Top-1 (\%) & Top-5 (\%) & $\Delta$Top-5 (\%)\\ \hline
ResNet50 + Common Training (CT) + Epoch 200 (E200)               & 76.70      & +0.00       & 93.15  & +0.00    \\
ResNet50 + Stimulative Training (ST) + Epoch 200 (E200)~\cite{ye2022stimulative}          & 77.56      & +0.86       & 93.85  & +0.70   \\
ResNet50 + ST + KL- + Snet6 + E200                                   & 78.53      & +1.83       & 94.22   & +1.07   \\
ResNet50 + ST + SITrans + E200                                     & 78.56      & +1.86       & 94.22  & +1.07    \\
ResNet50 + ST + KL- + Snet6 + SITrans + E200                           & 79.83      & +3.13       & 94.81   & +1.66   \\
ResNet50 + ST + KL- + Snet6 + SITrans + Vrs240 + E200                    & 79.91      & +3.21       & 94.95   & +1.80   \\
ResNet50 + ST + KL- + Snet6 + SITrans + Vrs240 + Tcs200 + E200             & 80.04      & +3.34       & 95.08   & +1.93   \\
ResNet50 + ST + KL- + Snet6 + SITrans + Vrs240 + Tcs200 + InterSSR + E200    & 80.50      & +3.80       & 95.26   & +2.11   \\
ResNet50 + ST + KL- + Snet6 + SITrans + Vrs240 + Tcs200 + InterSSR + UA + E400 & 80.97      & +4.27       & 95.34    & +2.19  \\ \hline
\end{tabular}
\vspace{-1mm}
\end{table*}

\begin{table}[]
\caption{Comparison with different techniques. All other results are taken from the cited paper. Our ST++ can achieve new SOTA performance without bells and whistles (no extra data or models are used, and no structures are changed).}
\vspace{-2mm}
\centering
\label{table:Comparison with different techniques}
\setlength\tabcolsep{4pt} 
\begin{tabular}{l|c|cc}
\hline
\multicolumn{1}{c|}{\multirow{2}{*}{Model}} & \multirow{2}{*}{FLOPs} & \multicolumn{2}{c}{Accuracy}                 \\ \cline{3-4} 
\multicolumn{1}{c|}{}                       &                        & \multicolumn{1}{c|}{Top-1 (\%)} & Top-5 (\%) \\ \hline
ResNet50~\cite{yuan2020revisiting}                                   & 4.1 G                  & \multicolumn{1}{c|}{75.77}      & -      \\ 
ResNet50~\cite{vryniotis2021train}                                   & 4.1 G                  & \multicolumn{1}{c|}{76.13}      & 92.86      \\ 
ResNet50~\cite{muller2021trivialaugment}                                   & 4.1 G                  & \multicolumn{1}{c|}{77.20}      & 93.43      \\ 
ResNet50~\cite{yun2019cutmix,cubuk2020randaugment,cubuk2019autoaugment}  & 4.1 G                  & \multicolumn{1}{c|}{76.32}      & 92.95      \\ \hline
ResNet50 + Cutout~\cite{devries2017improved}                          & 4.1 G                  & \multicolumn{1}{c|}{77.07}      & 93.34      \\
ResNet50 + Mixup~\cite{zhang2017mixup}                           & 4.1 G                  & \multicolumn{1}{c|}{77.90}      & 93.90      \\
ResNet50 + CutMix~\cite{yun2019cutmix}                          & 4.1 G                  & \multicolumn{1}{c|}{78.60}      & 94.08      \\
ResNet50 + Rand Erasing~\cite{zhong2020random}                    & 4.1 G                  & \multicolumn{1}{c|}{77.25}      & 93.31      \\ \hline
ResNet50 + StochDepth~\cite{huang2016deep}                      & 4.1 G                  & \multicolumn{1}{c|}{77.53}      & 93.73      \\
ResNet50 + Droppath~\cite{larsson2016fractalnet}                        & 4.1 G                  & \multicolumn{1}{c|}{77.10}      & 93.50      \\
ResNet50 + Dropblock~\cite{ghiasi2018dropblock}                       & 4.1 G                  & \multicolumn{1}{c|}{78.13}      & 94.02      \\
ResNet50 + ShakeDrop~\cite{yamada2018shakedrop}                       & 4.1 G                  & \multicolumn{1}{c|}{77.50}      & -          \\ \hline
ResNet50 + Rand Augment~\cite{cubuk2020randaugment}                    & 4.1 G                  & \multicolumn{1}{c|}{77.60}      & 93.80      \\
ResNet50 + Auto Augment~\cite{cubuk2019autoaugment}                    & 4.1 G                  & \multicolumn{1}{c|}{77.60}      & 93.80      \\
ResNet50 + Uniform Augment~\cite{lingchen2020uniformaugment}                 & 4.1 G                  & \multicolumn{1}{c|}{77.63}      & -          \\
ResNet50 + Trivial Augment~\cite{muller2021trivialaugment}                 & 4.1 G                  & \multicolumn{1}{c|}{78.07}      & 93.92      \\ \hline
ResNet50 + Repeated Aug~\cite{berman2019multigrain}                    & 4.1 G                  & \multicolumn{1}{c|}{76.86}      & -          \\
ResNet50 + Label Smoothing~\cite{yuan2020revisiting}                 & 4.1 G                  & \multicolumn{1}{c|}{76.38}      & -          \\
ResNet50 + GradAug~\cite{yang2020gradaug}                         & 4.1 G                  & \multicolumn{1}{c|}{78.79}      & 94.38      \\
ResNet50 + bag of tricks~\cite{he2019bag}                   & 4.3 G                  & \multicolumn{1}{c|}{79.29}      & 94.63      \\ \hline
ResNet50 + ST++ (A1)                       & 4.1 G                  & \multicolumn{1}{c|}{80.50}      & 95.26      \\
ResNet50 + ST++ (A2)                       & 4.1 G                  & \multicolumn{1}{c|}{80.97}      & 95.34      \\ \hline
\end{tabular}
\vspace{-1mm}
\end{table}

\begin{table*}[]
\caption{Influence of various training procedures that takes from different papers for ResNet50 on ImageNet. Note that all other results are taken from the Timm libary~\cite{wightman2021resnet} and PyTorch official version~\cite{vryniotis2021train}. Without using any other training techniques or tricks, our ST++ (A1) can perform better than most training procedures that combine plenty of ingredients with carefully-tuned hyper-parameters. Besides, our method is essentially complementary to other techniques. For example, with only uniform augment~\cite{lingchen2020uniformaugment}, our ST++ (A2) can already reach a new SOTA Top-1 accuracy of 81.0\%.
}
\vspace{-2mm}
\centering
\label{table:Various training procedures}
\setlength\tabcolsep{4pt} 
\begin{tabular}{l|cccc|ccc|ccc|cc}
\hline
\multicolumn{1}{c|}{\multirow{3}{*}{\begin{tabular}[c]{@{}c@{}}Procedure\\ Refenrence\end{tabular}}} & \multicolumn{4}{c|}{Previous approaches}                                                                 & \multicolumn{3}{c|}{Timm-best}                                                 & \multicolumn{3}{c|}{PyTorch}                                                 & \multicolumn{2}{c}{ST++}                        \\ \cline{2-13} 
\multicolumn{1}{c|}{}                                                                                & ResNet                 & FixRes                 & DeiT                   & FAMS (×4)                     & A1                       & A2                       & A3                       & A1                       & A2                       & A3                     & A1                     & A2                     \\
\multicolumn{1}{c|}{}                                                                                & {~\cite{he2016deep}}               & {~\cite{touvron2019fixing}}               & {~\cite{touvron2021training}}               & {~\cite{dollar2021fast}}                   & {~\cite{wightman2021resnet}}                  & {~\cite{wightman2021resnet}}                  & {~\cite{wightman2021resnet}}                  & {~\cite{vryniotis2021train}}                  & {~\cite{vryniotis2021train}}                  & {~\cite{vryniotis2021train}}                &                        &                        \\ \hline
Train Crop Size                                                                                            & 224                    & 224                    & 224                    & 224                           & 224                      & 224                      & 160                      & 176                      & 176                      & 224                    & 200                    & 200                    \\
Val Resize Size                                                                                      & 256                    & 256                    & 256                    & 256                           & 236                      & 236                      & 236                      & 232                      & 256                      & 256                    & 240                    & 240                    \\
Val Crop Size                                                                                       & 224                    & 224                    & 224                    & 224                           & 224                      & 224                      & 224                      & 224                      & 224                      & 224                    & 224                    & 224                    \\ \hline
Epochs                                                                                               & 90                     & 120                    & 300                    & 400                           & 600                      & 300                      & 100                      & 600                      & 600                      & 90                     & 200                    & 400                    \\
Batch size                                                                                           & 256                    & 512                    & 1024                   & 1024                          & 2048                     & 2048                     & 2048                     & 1024                     & 1024                     & 256                    & 512                    & 512                    \\
Optimizer                                                                                            & SGD-M                  & SGD-M                  & AdamW                  & SGD-M                         & LAMB                     & LAMB                     & LAMB                     & SGD-M                    & SGD-M                    & SGD-M                  & SGD-M                  & SGD-M                  \\
LR                                                                                                   & 0.1                    & 0.2                    & 10\textasciicircum{}-3 & 2                             & 5×10\textasciicircum{}-3 & 5×10\textasciicircum{}-3 & 8×10\textasciicircum{}-3 & 0.5                      & 0.5                      & 0.1                    & 0.2                    & 0.2                    \\
LR decay                                                                                             & step                   & step                   & cosine                 & step                          & cosine                   & cosine                   & cosine                   & cosine                   & cosine                   & step                   & cosine                 & cosine                 \\
Decay rate                                                                                           & 0.1                    & 0.1                    & ×                      & 0.02\textasciicircum{}(t/400) & ×                        & ×                        & ×                        & ×                        & ×                        & 0.1                    & ×                      & ×                      \\
Decay epochs                                                                                         & 30                     & 30                     & ×                      & 1                             & ×                        & ×                        & ×                        & ×                        & ×                        & 30                     & ×                      & ×                      \\
Weight decay                                                                                         & 10\textasciicircum{}-4 & 10\textasciicircum{}-4 & 0.05                   & 10\textasciicircum{}-4        & 0.01                     & 0.02                     & 0.02                     & 2*10\textasciicircum{}-5 & 2*10\textasciicircum{}-5 & 10\textasciicircum{}-4 & 10\textasciicircum{}-4 & 10\textasciicircum{}-4 \\
Norm decay                                                                                    & ×                      & ×                      & ×                      & ×                             & ×                        & ×                        & ×                        & 0.0                        & 0.0                        & ×                      & ×                      & ×                      \\
Warmup epochs                                                                                        & ×                      & ×                      & 5                      & 5                             & 5                        & 5                        & 5                        & 5                        & 5                        & ×                      & ×                      & ×                      \\ \hline
Label smoothing                                                                                      & ×                      & ×                      & 0.1                    & 0.1                           & 0.1                      & ×                        & ×                        & 0.1                      & 0.1                      & ×                      & ×                      & ×                      \\
Dropout                                                                                              & ×                      & ×                      & ×                      & ×                             & ×                        & ×                        & ×                        & ×                        & ×                        & ×                      & ×                      & ×                      \\
StochDepth                                                                                           & ×                      & ×                      & 0.1                    & ×                             & 0.05                     & 0.05                     & ×                        & ×                        & ×                        & ×                      & ×                      & ×                      \\
Repeated Aug                                                                                         & ×                      & \Checkmark                      & \Checkmark                      & ×                             & \Checkmark                        & \Checkmark                        & ×                        & ×                        & ×                        & ×                      & ×                      & ×                      \\
Gradient Clip.                                                                                       & ×                      & ×                      & ×                      & ×                             & ×                        & ×                        & ×                        & ×                        & ×                        & ×                      & ×                      & ×                      \\ \hline
H. flip                                                                                              & \Checkmark                      & \Checkmark                      & \Checkmark                      & \Checkmark                             & \Checkmark                        & \Checkmark                        & \Checkmark                        & \Checkmark                        & \Checkmark                        & \Checkmark                      & \Checkmark                      & \Checkmark                      \\
RRC                                                                                                  & ×                      & \Checkmark                      & \Checkmark                      & \Checkmark                             & \Checkmark                        & \Checkmark                        & \Checkmark                        & \Checkmark                        & \Checkmark                        & \Checkmark                      & \Checkmark                      & \Checkmark                      \\
Rand Augment                                                                                         & ×                      & ×                      & 9/0.5                  & ×                             & 7/0.5                    & 7/0.5                    & 6/0.5                    & ×                        & ×                        & ×                      & ×                      & ×                      \\
Auto Augment                                                                                         & ×                      & ×                      & ×                      & \Checkmark                             & ×                        & ×                        & ×                        & ×                        & ×                        & ×                      & ×                      & ×                      \\
Trivial Augment                                                                                      & ×                      & ×                      & ×                      & ×                             & ×                        & ×                        & ×                        & \Checkmark                        & \Checkmark                        & ×                      & ×                      & ×                      \\
Uniform Augment                                                                                      & ×                      & ×                      & ×                      & ×                             & ×                        & ×                        & ×                        & ×                        & ×                        & ×                      & ×                      & \Checkmark                      \\
Mixup alpha                                                                                          & ×                      & ×                      & 0.8                    & 0.2                           & 0.2                      & 0.1                      & 0.1                      & 0.2                      & 0.2                      & ×                      & ×                      & ×                      \\
Cutmix alpha                                                                                         & ×                      & ×                      & 1                      & ×                             & 1                        & 1                        & 1                        & 1                        & 1                        & ×                      & ×                      & ×                      \\
Erasing prob.                                                                                        & ×                      & ×                      & 0.25                   & ×                             & ×                        & ×                        & ×                        & 0.1                      & 0.1                      & ×                      & ×                      & ×                      \\
ColorJitter                                                                                          & ×                      & \Checkmark                      & ×                      & ×                             & ×                        & ×                        & ×                        & \Checkmark                        & \Checkmark                        & \Checkmark                      & \Checkmark                      & \Checkmark                      \\
PCA lighting                                                                                         & \Checkmark                      & ×                      & ×                      & ×                             & ×                        & ×                        & ×                        & ×                        & ×                        & ×                      & \Checkmark                      & \Checkmark                      \\ \hline
SWA                                                                                                  & ×                      & ×                      & ×                      & \Checkmark                             & ×                        & ×                        & ×                        & ×                        & ×                        & ×                      & ×                      & ×                      \\
EMA                                                                                                  & ×                      & ×                      & ×                      & ×                             & ×                        & ×                        & ×                        & \Checkmark                        & ×                        & ×                      & ×                      & ×                      \\ \hline
CE loss                                                                                              & \Checkmark                      & \Checkmark                      & \Checkmark                      & \Checkmark                             & ×                        & ×                        & ×                        & \Checkmark                        & \Checkmark                        & \Checkmark                      & \Checkmark                      & \Checkmark                      \\
BCE loss                                                                                             & ×                      & ×                      & ×                      & ×                             & \Checkmark                        & \Checkmark                        & \Checkmark                        & ×                        & ×                        & ×                      & ×                      & ×                      \\
Mixed precision                                                                                      & ×                      & ×                      & \Checkmark                      & \Checkmark                             & \Checkmark                        & \Checkmark                        & \Checkmark                        & ×                        & ×                        & ×                      & ×                      & ×                      \\ \hline
Top-1 acc.                                                                                           & 75.3\%                 & 77.0\%                 & 78.4\%                 & 79.5\%                        & 80.4\%                   & 79.8\%                   & 78.1\%                   & 80.7\%                   & 80.2\%                   & 76.1\%                 & 80.5\%                 & 81.0\%                     \\ \hline
\end{tabular}
\end{table*}

\section{Experimental Verification}
In this section, we conduct comprehensive experiments and comparisons to demonstrate the effectiveness and superiority of ST++. First, we perform plenty of experiments on ImageNet-1k, verifying the effectiveness of different components and the superiority of ST++ over various training ingredients and recipes. Second, we show the effectiveness of ST++ pretrained model on a total of eight downstream tasks, and we further find that ST++ can also be applied to the finetuning process of downstream tasks to improve their performance. Thirdly, we present more verification experiments, including its effectiveness on various models and datasets, and its performance under different training costs.
All these experiments highlight the potential of the proposed training scheme.
\subsection{Verification on ImageNet}
\subsubsection{Effectiveness of Different Components}
To verify the effectiveness of the proposed three strategies (i.e., KL-+Snet6, SITrans, and InterSST), we progressively add these components to the original ST and compare their main network performance. All experiments are performed with ResNet50 on ImageNet. As shown in Table~\ref{table:Effectiveness of three strategies}, original ST can already improve the Top-1 classification results of CT by 0.86\%. When adding KL-+Snet6 and SITrans to ST respectively, the Top-1 performance improvement increases to 1.83\% and 1.86\% respectively. When adding both KL-+Snet6 and SITrans to ST, the Top-1 performance improvement is up to 3.13\%. When we further adjust the validation resize size and train crop size, it can not only get a slightly higher performance gain (3.34\% vs 3.13\%) but also speed up the training process, similar to the conclusions in~\cite{touvron2019fixing}. Then, when we add the above strategies and InterSSR to ST, the Top-1 performance gain is boosted to 3.80\%. In addition, the proposed stimulative training and three simple-but-effective strategies are complementary to other training ingredients in essence. To verify this, on the basis of the above strategies, when we further add a data augmentation strategy called uniform augment~\cite{lingchen2020uniformaugment} and extend the training epoch to 400, the Top-1 performance gain continues to increase to 4.27\%. In summary, the proposed stimulative training and  its associated three strategies, although simple, can remarkably improve the performance of residual networks and serve as a strong booster for the current training scheme.

\subsubsection{Comparison with Various Training Ingredients}
To demonstrate the superiority of ST++, we compare it with many kinds of commonly-used training ingredients. These training ingredients can be roughly divided into four categories: 1) Individual data augmentation: Cutout~\cite{devries2017improved}, Mixup~\cite{zhang2017mixup}, CutMix~\cite{yun2019cutmix}, and Rand Erasing~\cite{zhong2020random}; 2) Network regularization technologies: StochDepth~\cite{huang2016deep}, Droppath~\cite{larsson2016fractalnet}, Dropblock~\cite{ghiasi2018dropblock}, and ShakeDrop~\cite{yamada2018shakedrop}; 3) Complex data augmentation: Rand Augment~\cite{cubuk2020randaugment}, Auto Augment~\cite{cubuk2019autoaugment}, Uniform Augment~\cite{lingchen2020uniformaugment}, and Trivial Augment~\cite{muller2021trivialaugment}; 4) Other representative methods: Repeated Augmentation~\cite{berman2019multigrain}, Label Smoothing~\cite{yuan2020revisiting}, GradAug~\cite{yang2020gradaug} and bag of tricks~\cite{he2019bag}. For the above training ingredients, we report their published best results for ResNet50 on ImageNet for a fair comparison. For ST++, we design two training procedures, namely ST++(A1) and ST++(A2), where the former does not use any of the above ingredients, and the latter combines the Uniform Augment strategy. For ResNet50, we report several different baselines used in different papers.
All the results are shown in Table~\ref{table:Comparison with different techniques}. 
As we can see, with different training ingredients, different performances are obtained, ranging from 76.38\% to 79.29\%. 
As a comparison, ST++(A1) firstly achieves more than 80\% of Top-1 accuracy (up to 80.50\%), without bells and whistles (no extra data or models are used, and no structures are changed). Further, ST++(A2) that combines Uniform Augment can achieve even better results (i.e., 81.0\%), which means ST++ can be applied with other training ingredients to construct stronger training recipes.
\subsubsection{Comparison with Various Training Recipes}
On the one hand, we show that aligning the training settings and providing the training details when conducting experiments and comparisons is important. As demonstrated in Table~\ref{table:Comparison with different techniques}, for the commonly-used ResNet50, the baseline performances used in different papers may also be different, making it difficult to compare fairly between papers. As illustrated in Table~\ref{table:Various training procedures}, when adopting different training procedures taken from different papers such as ResNet~\cite{he2016deep}, FixRes~\cite{touvron2019fixing}, and DeiT~\cite{touvron2021training}, the performance of ResNet50 could vary widely. In this paper, we verify the effectiveness of ST++ under the same training settings in Table~\ref{table:Effectiveness of three strategies}, and provide all the training details of ST++ in Table~\ref{table:Various training procedures}.

On the other hand, to further show the superiority of ST++, we compare it with SOTA training procedures given by the Timm libary~\cite{wightman2021resnet} and PyTorch official~\cite{vryniotis2021train}, which combine plenty of ingredients with carefully-tuned hyper-parameters. All experiments are performed with ResNet50 on ImageNet, and the results are shown in Table~\ref{table:Various training procedures}. Without using any other training techniques or tricks, our ST++(A1) can achieve a high accuracy of 80.5\% and perform better than most training procedures that combine plenty of ingredients with carefully-tuned hyper-parameters. With only uniform augment~\cite{lingchen2020uniformaugment}, our ST++(A2) can already reach a new SOTA Top-1 accuracy of 81.0\%. Since our method is complementary to other ingredients, higher performance may be achieved by combining more other ingredients.

\begin{table*}[]
\caption{Verification on object detection. We choose four classical detection frameworks, including Faster R-CNN~\cite{girshick2015fast}, Cascade R-CNN~\cite{cai2018cascade}, FCOS~\cite{tian2019fcos}, and ATSS~\cite{zhang2020bridging}, and train each framework by schedule 1$\times$ using AdamW on COCO.}
\vspace{-2mm}
\centering
\label{table:object detection AdamW}
\begin{tabular}{c|c|c|c|c|c|c|c|c}
\hline
Backbone & Img Top1   & Optim & lr     & WD   & \begin{tabular}[c]{@{}c@{}}Faster R-CNN\\    (Bbox mAP)\end{tabular} & \begin{tabular}[c]{@{}c@{}}Cascade R-CNN\\ (Bbox mAP)\end{tabular} & \begin{tabular}[c]{@{}c@{}}FCOS\\ (Bbox mAP)\end{tabular} & \begin{tabular}[c]{@{}c@{}}ATSS\\ (Bbox mAP)\end{tabular} \\ \hline
R50-base & 76.1       & AdamW & 0.0001 & 0.10 & 38.1                                                                & 41.4                                                               & 36.0                                                      & 39.4                                                      \\
R50-timm & 80.4 & AdamW & 0.0001 & 0.10 & 38.2(\textcolor{red}{+0.1})                                                          & 42.1(\textcolor{red}{+0.7})                                                         & 33.8(\textcolor{green}{-2.2})                                                & 39.4(0.0)                                                 \\
R50-tnr  & 80.7 & AdamW & 0.0001 & 0.10 & 40.0(\textcolor{red}{+1.9})                                                          & 43.3(\textcolor{red}{+1.9})                                                         & 30.8(\textcolor{green}{-5.2})                                                & 41.6(\textcolor{red}{+2.2})                                                \\ \hline
R50-ST   & 81.0 & AdamW & 0.0001 & 0.10 & 40.0(\textcolor{red}{+1.9})                                                          & 43.4(\textcolor{red}{+2.0})                                                         & 38.1(\textcolor{red}{+2.1})                                                & 41.7(\textcolor{red}{+2.3})                                                \\
R50-ST*  & 81.0 & AdamW & 0.0001 & 0.10 & 41.9(\textcolor{red}{+3.8})                                                          & 45.1(\textcolor{red}{+3.7})                                                         & 40.7(\textcolor{red}{+4.7})                                                & 43.6(\textcolor{red}{+4.2})                                                \\ \hline
\end{tabular}
\end{table*}

\begin{table*}[]
\caption{Verification on object detection. We choose four classical detection frameworks, including Faster R-CNN~\cite{girshick2015fast}, Cascade R-CNN~\cite{cai2018cascade}, FCOS~\cite{tian2019fcos}, and ATSS~\cite{zhang2020bridging}, train each framework by schedule 1$\times$ using SGD on COCO.}
\vspace{-2mm}
\centering
\label{table:object detection SGD}
\begin{tabular}{c|c|c|c|c|c|c|c|c}
\hline
Backbone & Img Top1 & Optim & lr     & WD   & \begin{tabular}[c]{@{}c@{}}Faster R-CNN\\    (Bbox mAP)\end{tabular} & \begin{tabular}[c]{@{}c@{}}Cascade R-CNN\\ (Bbox mAP)\end{tabular} & \begin{tabular}[c]{@{}c@{}}FCOS\\ (Bbox mAP)\end{tabular} & \begin{tabular}[c]{@{}c@{}}ATSS\\ (Bbox mAP)\end{tabular} \\ \hline
R50-base & 76.1     & SGD   & 0.0001 & 0.10 & 37.2                                                                & 40.2                                                               & 36.3                                                      & 39.3                                                      \\
R50-timm & 80.4     & SGD   & 0.0001 & 0.10 & 15.0(\textcolor{green}{-22.2})                                                         & 20.5(\textcolor{green}{-19.7})                                                        & 16.9(\textcolor{green}{-19.4})                                               & 24.9(\textcolor{green}{-14.4})                                               \\
R50-tnr  & 80.7     & SGD   & 0.0001 & 0.10 & 28.4(\textcolor{green}{-8.8})                                                          & 34.8(\textcolor{green}{-5.4})                                                         & 30.1(\textcolor{green}{-6.2})                                                & 36.0(\textcolor{green}{-3.3})                                                \\ \hline
R50-ST   & 81.0     & SGD   & 0.0001 & 0.10 & 38.3(\textcolor{red}{+1.1})                                                          & 41.3(\textcolor{red}{+1.1})                                                         & 37.7(\textcolor{red}{+1.4})                                                & 41.0(\textcolor{red}{+1.7})                                                \\
R50-ST*  & 81.0     & SGD   & 0.0001 & 0.10 & 40.1(\textcolor{red}{+2.9})                                                          & 43.3(\textcolor{red}{+3.1})                                                         & 38.7(\textcolor{red}{+2.4})                                                & 42.6(\textcolor{red}{+3.3})                                                \\ \hline
\end{tabular}
\end{table*}

\begin{table*}[]
\caption{Verification on panoptic segmentation. We train Panoptic FPN~\cite{kirillov2019panoptic} by schedule 1$\times$ using AdamW on COCO.}
\vspace{-2mm}
\centering
\label{table:panoptic segmentation AdamW}
\begin{tabular}{c|c|c|c|c|c|c|c}
\hline
Backbone & Img Top1 & Optim & lr     & WD   & PQ             & SQ             & RQ             \\ \hline
R50-base & 76.1     & AdamW & 0.0001 & 0.10 & 41.834         & 78.213         & 51.104         \\
R50-timm & 80.4     & AdamW & 0.0001 & 0.10 & 40.937(\textcolor{green}{-0.897}) & 78.188(\textcolor{green}{-0.025}) & 50.170(\textcolor{green}{-0.934}) \\
R50-tnr  & 80.7     & AdamW & 0.0001 & 0.10 & 42.265(\textcolor{red}{+0.431}) & 78.871(\textcolor{red}{+0.658}) & 51.753(\textcolor{red}{+0.649}) \\ \hline
R50-ST   & 81.0     & AdamW & 0.0001 & 0.10 & 42.489(\textcolor{red}{+0.655}) & 78.927(\textcolor{red}{+0.714}) & 51.823(\textcolor{red}{+0.719}) \\
R50-ST*  & 81.0     & AdamW & 0.0001 & 0.10 
& 43.517(\textcolor{red}{+1.683})
& 78.956(\textcolor{red}{+0.743})
& 53.019(\textcolor{red}{+1.915})
            \\ \hline
\end{tabular}
\end{table*}

\begin{table*}[]
\caption{Verification on panoptic segmentation. We train Panoptic FPN~\cite{kirillov2019panoptic} by schedule 1$\times$ using SGD on COCO.}
\vspace{-2mm}
\centering
\label{table:panoptic segmentation SGD}
\begin{tabular}{c|c|c|c|c|c|c|c}
\hline
Backbone & Img Top1 & Optim & lr   & WD     & PQ              & SQ             & RQ              \\ \hline
R50-base & 76.1     & SGD   & 0.02 & 0.0001 & 39.821          & 77.739         & 48.736          \\
R50-timm & 80.4     & SGD   & 0.02 & 0.0001 & 24.634(\textcolor{green}{-15.187}) & 68.506(\textcolor{green}{-9.233}) & 31.101(\textcolor{green}{-17.635}) \\
R50-tnr  & 80.7     & SGD   & 0.02 & 0.0001 & 35.513(\textcolor{green}{-4.308})  & 76.253(\textcolor{green}{-1.486}) & 43.748(\textcolor{green}{-4.988})  \\ \hline
R50-ST   & 81.0     & SGD   & 0.02 & 0.0001 & 40.770(\textcolor{red}{+0.949})  & 77.849(\textcolor{red}{+0.110})  & 49.809(\textcolor{red}{+1.073})  \\
R50-ST*  & 81.0     & SGD   & 0.02 & 0.0001 & 41.344(\textcolor{red}{+1.523})  & 78.375(\textcolor{red}{+0.636}) &  50.566(\textcolor{red}{+1.830})  \\ \hline
\end{tabular}
\end{table*}

\begin{table}[]
\caption{Verification on instance segmentation. We train MaskRCNN~\cite{he2017mask} by schedule 1$\times$ using AdamW on COCO.}
\vspace{-2mm}
\centering
\label{table:instance segmentation AdamW}
\scriptsize
\setlength\tabcolsep{4pt} 
\begin{tabular}{c|c|c|c|c|c|c}
\hline
Backbone & Img Top1 & Optim & lr     & WD   & Det mAP    & Seg mAP    \\ \hline
R50-base & 76.1     & AdamW & 0.0001 & 0.10 & 38.9       & 35.8       \\
R50-timm & 80.4     & AdamW & 0.0001 & 0.10 & 38.4(\textcolor{green}{-0.5}) & 35.9(\textcolor{red}{+0.1}) \\
R50-tnr  & 80.7     & AdamW & 0.0001 & 0.10 & 40.5(\textcolor{red}{+1.6}) & 37.1(\textcolor{red}{+1.3}) \\ \hline
R50-ST   & 81.0     & AdamW & 0.0001 & 0.10 & 40.7(\textcolor{red}{+1.8}) & 37.1(\textcolor{red}{+1.3}) \\
R50-ST*  & 81.0     & AdamW & 0.0001 & 0.10 & 42.5(\textcolor{red}{+3.6}) & 38.6(\textcolor{red}{+2.8}) \\ \hline
\end{tabular}
\end{table}

\begin{table}[]
\caption{Verification on instance segmentation. We train MaskRCNN~\cite{he2017mask} by schedule 1$\times$ using SGD on COCO.}
\vspace{-2mm}
\centering
\label{table:instance segmentation SGD}
\scriptsize
\setlength\tabcolsep{4pt} 
\begin{tabular}{c|c|c|c|c|c|c}
\hline
Backbone & Img Top1 & Optim & lr   & WD     & Det mAP     & Seg mAP     \\ \hline
R50-base & 76.1     & SGD   & 0.02 & 0.0001 & 37.8        & 34.5        \\
R50-timm & 80.4     & SGD   & 0.02 & 0.0001 & 17.5(\textcolor{green}{-20.3}) & 17.0(\textcolor{green}{-17.5}) \\
R50-tnr  & 80.7     & SGD   & 0.02 & 0.0001 & 30.6(\textcolor{green}{-7.2})  & 28.3(\textcolor{green}{-6.2})  \\ \hline
R50-ST   & 81.0     & SGD   & 0.02 & 0.0001 & 39.0(\textcolor{red}{+1.2})  & 35.4(\textcolor{red}{+0.9})  \\
R50-ST*  & 81.0     & SGD   & 0.02 & 0.0001 & 40.6(\textcolor{red}{+2.8})  & 37.0(\textcolor{red}{+2.5})  \\ \hline
\end{tabular}
\end{table}

\begin{table}[]
\caption{Verification on semantic segmentation. We train FPN R50~\cite{kirillov2019panoptic} 
using 512$\times$1024 resolution on Cityscapes.}
\vspace{-2mm}
\centering
\label{table:semantic segmentation}
\scriptsize
\setlength\tabcolsep{3pt} 
\begin{tabular}{c|c|c|c|c|c|c}
\hline
Backbone & Img Top1 & Optim & lr     & WD            & mIoU         & mAcc         \\ \hline
R50-base & 76.1     & AdamW & 0.0005 & 0.05         & 76.38        & 83.89        \\
R50-timm & 80.4     & AdamW & 0.0005 & 0.05 & 76.77(\textcolor{red}{+0.39}) & 83.87(\textcolor{green}{-0.02}) \\
R50-tnr  & 80.7     & AdamW & 0.0005 & 0.05 & 76.14(\textcolor{green}{-0.24}) & 83.36(\textcolor{green}{-0.53}) \\ \hline
R50-ST   & 81.0     & AdamW & 0.0005 & 0.05 &  77.58(\textcolor{red}{+1.20}) & 84.95(\textcolor{red}{+1.06}) \\
\hline
\end{tabular}
\end{table}

\begin{table}[]
\caption{Verification on pose estimation. We train SimpleBaseline2D~\cite{xiao2018simple} 
using 256$\times$192 resolution on COCO.}
\vspace{-2mm}
\centering
\label{table:pose estimation}
\scriptsize
\setlength\tabcolsep{3pt} 
\begin{tabular}{c|c|c|c|c|c|c}
\hline
Backbone & Img Top1 & Optim & lr     & WD & AP              & AR              \\ \hline
R50-base & 76.1     & Adam  & 0.0005 & -  & 0.7173          & 0.7727          \\
R50-timm & 80.4     & Adam  & 0.0005 & -  & 0.6952(\textcolor{green}{-0.0221}) & 0.7522(\textcolor{green}{-0.0205}) \\
R50-tnr  & 80.7     & Adam  & 0.0005 & -  & 0.7199(\textcolor{red}{+0.0026}) & 0.7755(\textcolor{red}{+0.0028}) \\ \hline
R50-ST   & 81.0     & Adam  & 0.0005 & -  & 0.7223(\textcolor{red}{+0.0050}) & 0.7777(\textcolor{red}{+0.0050}) \\
\hline
\end{tabular}
\end{table}

\subsection{Verification on Various Downstream Tasks}
\label{section:downstream tasks}
In this section, we aim to investigate the effectiveness of  stimulative training on various downstream tasks, e.g., object detection, panoptic segmentation, instance segmentation, semantic segmentation, and pose estimation. For comprehensive verification, we choose eight well-known frameworks including Faster R-CNN~\cite{girshick2015fast}, Cascade R-CNN~\cite{cai2018cascade}, FCOS~\cite{tian2019fcos}, ATSS~\cite{zhang2020bridging}, Panoptic FPN~\cite{kirillov2019panoptic}, MaskRCNN~\cite{he2017mask}, FPN R50~\cite{kirillov2019panoptic}, and SimpleBaseline2D~\cite{xiao2018simple}, and employ two different optimization strategies. All these frameworks are implemented through commonly-used toolboxes, namely, MMDetection~\cite{chen2019mmdetection}, MMSegmentation~\cite{mmseg2020}, and MMPose~\cite{mmpose2020}. For an intuitive and fair comparison, we also evaluate the ResNet50 backbone pretrained with the ResNet recipe~\cite{he2016deep}, the Timm-best A1 recipe~\cite{wightman2021resnet}, and the PyTorch A1 recipe~\cite{vryniotis2021train} on these frameworks.
\subsubsection{Effectiveness of ST++ Pretrained Model}
We first verify the effectiveness of ST++ pretrained model (i.e., R50-ST) on different kinds of downstream tasks. The results are shown in Table~\ref{table:object detection AdamW}, Table~\ref{table:object detection SGD}, Table~\ref{table:panoptic segmentation AdamW}, Table~\ref{table:panoptic segmentation SGD}, Table~\ref{table:instance segmentation AdamW}, Table~\ref{table:instance segmentation SGD}, Table~\ref{table:semantic segmentation}, Table~\ref{table:pose estimation}. As we can see, when using the AdamW optimizer, compared to the ResNet recipe~\cite{he2016deep} pretrained model (i.e., R50-base), Timm-best A1 recipe~\cite{wightman2021resnet} pretrained model (i.e., R50-timm) and PyTorch A1 recipe~\cite{vryniotis2021train} pretrained model (i.e., R50-tnr) can improve the performance of most frameworks while still suffering from the degraded results on some frameworks such as FCOS, Panoptic FPN, and FPN R50. When using the SGD optimizer, compared to R50-base, R50-timm, and R50-tnr lead to significantly lower performances for all frameworks. The reason may be that Timm-best A1 recipe and PyTorch A1 recipe both employ plenty of training ingredients with carefully-tuned hyper-parameters. As a comparison, our method R50-ST that uses few tricks can achieve consistent significant performance gains across different frameworks and optimizers. For example, when using the AdamW optimizer, the average performance gain for Faster R-CNN, Cascade R-CNN, FCOS, ATSS detection frameworks is up to about 2.1\% Bbox mAP.

\subsubsection{ST++ on Downstream Tasks Finetuning}
We then verify the effectiveness of ST++ on the finetuning process (i.e., R50-ST*) of downstream tasks. In detail, after loading ST++ pretrained model, at each finetuning iteration, we sample subnetworks following the inner-stage and inter-stage sampling rules and provide subnetworks with random smaller inputs. For different downstream tasks, we always train subnetworks with the ground truth for simplicity. The results are shown in Table~\ref{table:object detection AdamW}, Table~\ref{table:object detection SGD}, Table~\ref{table:panoptic segmentation AdamW}, Table~\ref{table:panoptic segmentation SGD}, Table~\ref{table:instance segmentation AdamW}, Table~\ref{table:instance segmentation SGD}. As we can see, under different frameworks and optimizers, R50-ST* can always remarkably improve the performance on the basis of R50-ST. For example, when using the AdamW optimizer, the average performance for Faster R-CNN, Cascade R-CNN, FCOS, ATSS detection frameworks is about 4.1\% Bbox mAP higher than R50-base and 2.0\% Bbox mAP higher than R50-ST. Such results further verify the generality of stimulative training.

\subsection{More Experiments and Findings}
\subsubsection{Effectiveness on Various Models and Datasets}
We further compare the performance of stimulative training and common training on various models and datasets. We show some implementation details below. For MobileNetV3 on CIFAR10, MobileNetV3 on CIFAR100, and ResNet50 on CIFAR100, we always use the SGD optimizer and train the model for 500 epochs with a batch size of 64. For ResNet101 on ImageNet, we use the SGD optimizer and train the model for 100 epochs with a batch size of 512. For VIT on ImageNet, we use the AdamW optimizer and train the model for 100 epochs with a batch size of 2048. More details can be found in the Appendix.

\begin{figure}[t]
\centering
\includegraphics[width=0.4\textwidth]{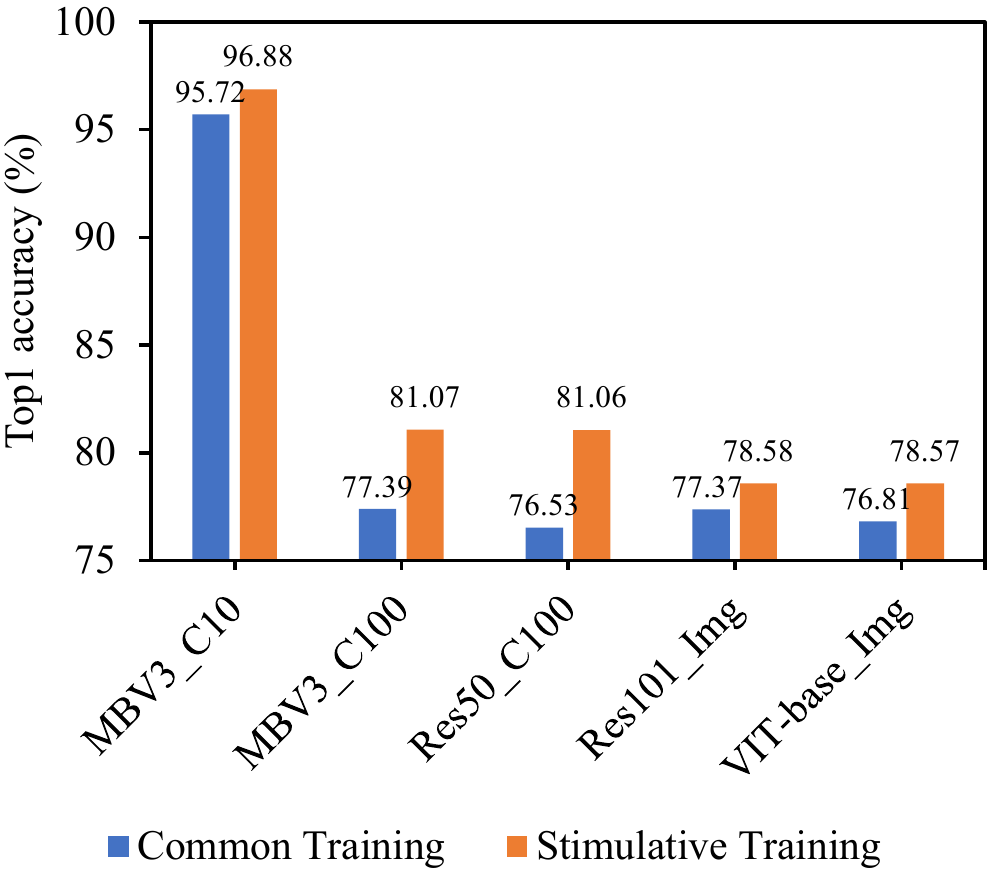}
\vspace{-2mm}
\caption{Comparison between our stimulative training and common training on various models and datasets.}
\label{fig:various_models}
\end{figure}

\begin{figure}[t]
\centering
\includegraphics[width=0.4\textwidth]{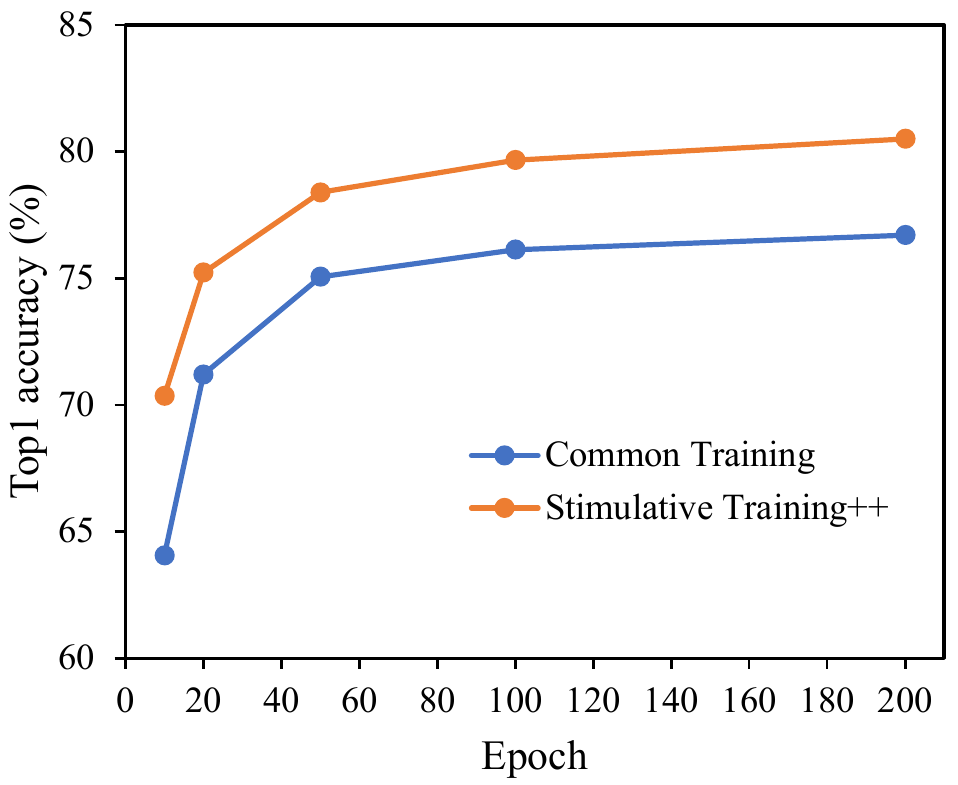}
\vspace{-2mm}
\caption{Comparison between our stimulative training and common training when using different training epochs.
}
\label{fig:less_epochs}
\end{figure}

Fig.~\ref{fig:various_models} shows the comparison between common training and stimulative training.
It is obvious that stimulative training can noticeably improve the performance of various networks on various datasets compared to common training (e.g., by 4.53\% for ResNet50 on CIFAR-100, by 3.68\% for MobileNetV3 on CIFAR-100, and by 1.76\% for VIT on ImageNet). On one hand, such results validate the importance of solving the network loafing problem, since we can improve the performance of a given residual network by training its subnetworks. On the other hand, experimental results on various models and datasets validate the generalization of stimulative training, thus we advocate it as a general technology for training residual networks. 

\subsubsection{Higher performance with Fewer Epochs}
As shown in Table~\ref{table:Various training procedures}, we find that our stimulative training can reach higher performance with fewer epochs. For example, ST++ A1 can reach 80.5\% Top1 accuracy with only 200 epochs, while Timm-best A1 needs 600 epochs to reach 80.4\% Top1 accuracy. In order to further validate this, we compare the performance of stimulative training and common training when using different training epochs. As shown in Fig.~\ref{fig:less_epochs}, the performance of stimulative training++ with only 50 epochs is already much higher than that of common training with 200 epochs, which further verifies that stimulative training can reach higher performance with fewer epochs. This property will be very useful in settings where reading and writing are very time-consuming.

\section{Theoretical Analysis}
\textbf{Analysis 1: Why the performance of subnetworks can be improved.}
Based on Eq.~\ref{fomulation:ut loss}, our method can be treated as optimizing $CE(\mathcal{Z}(\theta_{D_m},x ),y)$ and $KL(\mathcal{Z}(\theta_{D_m},x),\mathcal{Z}(\theta_{D_s} ,x))$ at the same time. According to the convergence of stochastic gradient descent, after enough iterations, both losses can be bounded by a constant, which is formulated as
\begin{align}
&CE(\mathcal{Z}(\theta_{D_m},x ),y)=-\sum_{i=1}^{N} y_i \log p_{i}^m < \epsilon_1 \\
&KL(\mathcal{Z}(\theta_{D_m},x),\mathcal{Z}(\theta_{D_s} ,x))=\mathbb{E}_\Theta\left [  \sum_{i=1}^{N} p_{i}^m \log \frac{p_{i}^m}{p_{i}^s} \right ] < \epsilon_2  
\end{align}
where $N$ is the number of categories. The $\epsilon_1$ and $\epsilon_2$ are the tiny constants for corresponding losses. $\Theta$ is the set of all subnetworks. The $p^m$ and $p^s$ are the classification probability of the main network and subnetwork, respectively. Furthermore, we can prove that \textit{the absolute difference of cross entropy of subnetworks and main network can be bounded by a constant} (detailed proof can be found in the Appendix), which can be formulated as
\begin{small}
\begin{align}
\label{formulation:ce gap} \left |   CE(\mathcal{Z}(\theta_{D_m},x ),y)- \mathbb E _\Theta \left [  CE(\mathcal{Z}(\theta_{D_s},x ),y)\right ]\right | 
<\frac{\epsilon_2 + \log N }{e^{-\epsilon_1}} + \epsilon_1 
\end{align}
\end{small}
\noindent where $\frac{\epsilon_2 + \log N }{e^{-\epsilon_1}} + \epsilon_1$ is a small constant after optimization. Eq.~\ref{formulation:ce gap} demonstrates that the performance gap between subnetworks and the main network is constrained. Thus, \textit{as the rising of the main network performance, subnetworks performance will be improved step by step}. This conclusion still holds true under KL- loss, since we remove the influence of logits amplitude before softmax activation.

\textbf{Analysis 2: Why the performance of given residual networks can be improved.} Neural network training is in fact a form of polynomial regression, and any neural network with activation functions may roughly correspond to a fitted polynomial~\cite{cheng2018polynomial}. For simplifying notations, suppose that both input $x$ and output $y$ in a neural network have only one element. Then, we can consider the training of a neural network as the fitting of a polynomial function.
\begin{align}
\label{fomulation:polynomial} 
y=F(x)=c_{0}+c_{1}x+c_{2}x^{2}+\cdots+c_{n} x^{n}+\ldots
\end{align}
where $c_{1}, \ c_{2},\ \ldots,\ c_{n},\ \ldots$ represent the polynomial coefficients. Based on training samples, we can establish the polynomial (neural network), which denotes the mapping relationship between inputs and outputs. For a test sample $x$, there always exists a point $x_{0}$ (e.g., a training sample) close enough to $x$, which can obtain an accurate output via the trained model. According to Taylor expansion, $F(x)$ at $x_{0}$ can be computed as :
\begin{align}
\label{fomulation:taylor} 
\begin{array}{l}
y=F(x)=\frac{F(x_{0})}{0 !}+\frac{F^{\prime}(x_{0})}{1 !}(x-x_{0})+ \\
\frac{F^{\prime \prime}(x_{0})}{2 !}(x-x_{0})^{2}+\cdots+\frac{F^{(n)}(x_{0})}{n !}(x-x_{0})^{n}+R_{n}(x)
\end{array}
\end{align}
where $R_{n}(x)$ denotes the higher degree infinitesimal of $(x-x_{0})^{n}$. According to Eq.~\ref{fomulation:polynomial} and Eq.~\ref{fomulation:taylor}, it is obvious that different terms of the fitted polynomial have different effects, and the low-degree terms tend to have greater effects on the predicted performance. Based on this point, ~\cite{sun2022low} further figures out that shallow subnetworks in residual networks roughly correspond to low-degree polynomials consisting of low-degree terms, while deep subnetworks are opposite. Following this theory, the proposed stimulative training can not only \textit{strengthen the learning of different items} of the polynomial (neural network), but also \textit{pay more attention to the learning of low-degree items}. As a result, the final performance of given residual networks can be improved by stimulative training.

\section{Conclusions}
In this paper, we intend to understand and improve residual networks from a social psychology perspective of loafing. Based on the novel view that a residual network behaves like an ensemble network, we find that different kinds of residual networks invariably exhibit loafing-like behaviors that are consistent with the social loafing problem of social psychology. We define this previously overlooked problem as network loafing.
As the loafing problem hinders the productivity of each individual and the whole collective, we learn from social psychology and propose a stimulative training strategy to solve this problem. Specifically, we randomly sample a subnetwork and calculate the KL divergence loss between the sampled subnetwork and the given residual network for extra supervision. To further unleash the potential of stimulative training, we then propose three simple-yet-effective strategies, including KL- loss that only aligns the network logits direction, random smaller inputs for subnetworks, and inter-stage sampling rules.

Comprehensive experiments and analysis demonstrate the effectiveness of stimulative training as well as its three improved strategies.
For example, without using any extra data, model, trick, or changing the structure, the proposed method can improve the performance of ResNet50 on ImageNet to 80.5\% Top1 accuracy. With only uniform augmentation, the performance can be further improved to 81.0\% Top1 accuracy, better than the best training recipes provided by Timm library and PyTorch official version. 
In addition, we also validate the effectiveness of the proposed method on various datasets and networks and theoretically show why the performance of a given residual network and its subnetworks can be improved.

\section{Limitations}
As the first investigation of the network loafing problem, the proposed method is a positive pioneer-like exploration. We believe designing a more efficient method to solve the network loafing problem is a worthy research direction in the future. Since the residual structure is widely applied in numerous different types of models, including DenseNet, Transformer, CNN-Transformer combined network, MLP network, and Graph network. It will be of vital value to 
explore some proper methods to solve the loafing problem in these models. 
Moreover, the proposed method adopts main network logits as supervision to alleviate loafing, and analogously, the proposed method may be applied in self-supervised learning, e.g., we could explore incorporating stimulating training into the MAE pretraining procedure.

We believe that taking full advantage of splendid achievements in interdisciplinary research can help promote the development of deep learning. We hope that this paper can provide a new perspective to inspire more researchers to comprehend and improve deep neural networks from other fields such as social psychology.

\ifCLASSOPTIONcompsoc
  \section*{Acknowledgments}
\else
  \section*{Acknowledgment}
\fi
This work is supported by National Natural Science Foundation of China (No. 62071127, U1909207 ),  Shanghai Natural Science Foundation (No. 23ZR1402900), and Zhejiang Lab Project (No. 2021KH0AB05). Wanli Ouyang is supported by the Australian Research Council Grant DP200103223, Australian Medical Research Future Fund MRFAI000085, CRC-P Smart Material Recovery Facility (SMRF) – Curby Soft Plastics, and CRC-P ARIA - bionic visual-spatial prosthesis for the blind.

\ifCLASSOPTIONcaptionsoff
  \newpage
\fi



\bibliographystyle{IEEEtran}
\normalem
\bibliography{main_paper}
\end{document}